\documentclass[final,5p,times,twocolumn,authoryear]{elsarticle}

\usepackage{amssymb}
\usepackage{amsmath}

\usepackage{tikz}
\usepackage{pgfplots}
\pgfplotsset{compat=1.18}
\usepackage[caption=false,font=footnotesize,labelfont=rm,textfont=rm]{subfig}
\usepackage{caption}
\usepackage{multirow}
\usepackage{booktabs}
\usepackage{makecell}
\usepackage{booktabs}
\DeclareCaptionFormat{cont}{#1 (cont.)#2#3\par}
\journal{ISPRS Journal of Photogrammetry and Remote Sensing}
\usepackage{url}
\usepackage{hyperref}
\hypersetup{
    colorlinks=true,
    linkcolor=blue,
    filecolor=magenta,      
    urlcolor=cyan,
    pdftitle={RSTeller},
    pdfpagemode=FullScreen,
    }
\usepackage[capitalise]{cleveref}
\makeatletter
\renewcommand{\fnum@figure}{Fig. \thefigure}
\makeatother

\usepackage{microtype}
\usepackage{listings}
\usepackage{changes}
\usepackage{xcolor}
\usepackage[normalem]{ulem}

\definecolor{addcolor}{RGB}{255,0,255}    
\definecolor{delcolor}{RGB}{255,165,0}   
\definecolor{replcolor}{RGB}{138,43,226} 

\makeatletter
\AddToHook{cmd/added/before}{\def\Changes@AuthorColor{addcolor}}
\AddToHook{cmd/deleted/before}{\def\Changes@AuthorColor{delcolor}}
\AddToHook{cmd/replaced/before}{\def\Changes@AuthorColor{replcolor}}
\makeatother

\begin{document}


\begin{frontmatter}



\title{RSTeller: Scaling Up Visual Language Modeling in Remote Sensing with Rich Linguistic Semantics from Openly Available Data and Large Language Models}
\author{Junyao Ge}
\author{Xu Zhang}
\author{Yang Zheng}
\author{Kaitai Guo}
\author{Jimin Liang\corref{cor1}} 
\ead{jimleung@mail.xidian.edu.cn}
\cortext[cor1]{Corresponding author.}


\address{School of Electronic Engineering, Xidian University, Xi'an, Shaanxi 710071, China}

\begin{abstract}
Abundant, well-annotated multimodal data in remote sensing are pivotal for aligning complex visual remote sensing (RS) scenes with human language, enabling the development of specialized vision language models across diverse RS interpretation tasks. However, annotating RS images with rich linguistic semantics at scale demands expertise in RS and substantial human labor, making it costly and often impractical. In this study, we propose a workflow that leverages large language models (LLMs) to generate multimodal datasets with semantically rich captions at scale from plain OpenStreetMap (OSM) data for images sourced from the Google Earth Engine (GEE) platform. This approach facilitates the generation of paired remote sensing data and can be readily scaled up using openly available data. Within this framework, we present RSTeller, a multimodal dataset comprising over 1.3 million RS images, each accompanied by two descriptive captions. Extensive experiments demonstrate that RSTeller enhances the performance of multiple existing vision language models for RS scene understanding through continual pre-training. Our methodology significantly reduces the manual effort and expertise needed for annotating remote sensing imagery while democratizing access to high-quality annotated data. This advancement fosters progress in visual language modeling and encourages broader participation in remote sensing research and applications. The RSTeller dataset is available at \url{https://github.com/SlytherinGe/RSTeller}.

\end{abstract}




\end{frontmatter}




\section{Introduction}
\label{sec:intro}

Vision language models (VLMs) hold great promise in remote sensing (RS) as they offer a means to interpret complex visual scenes for non-expert users in a human-like manner and support the development of diverse RS applications. However, the effective training of VLMs hinges on the availability of large quantities of paired data to ensure the alignment of visual and linguistic semantics for optimal performance. The substantial data requirements present a significant challenge in scaling up VLMs for RS, impeding their broader applicability and performance enhancement. Consequently, there is a pressing need for automated, scalable, and efficient methods to generate datasets with paired RS imagery and semantic-rich captions. Such methods are essential for overcoming the bottleneck associated with data scarcity and facilitating the development of robust VLMs tailored for RS tasks. Moreover, the availability of an out-of-the-box, well-annotated dataset comprising RS images with rich linguistic semantics holds immense value. It not only streamlines the initial training of VLMs but also contributes to reducing the environmental impact associated with the individual generation of such data, thereby addressing sustainability concerns. 


\subsection{Related Works}



The intersection of computer vision and natural language processing (NLP) has witnessed remarkable advancements in recent years, propelled by the emergence of large-scale pre-trained models. When CLIP \citep{radford2021learning} was initially introduced, it swiftly captivated the research community with its remarkable performance in zero-shot classification and cross-modal retrieval tasks, rivaling some state-of-the-art models trained with full supervision. CLIP achieves this feat by aligning image and text representations through a contrastive learning strategy, endowing it with the capability for zero-shot inference using the learned representations. Inspired by this paradigm, the RS community embarked on leveraging similar strategies to construct VLMs tailored for RS tasks. Notable efforts such as those by \cite{liu2024remoteclip}, \cite{zhang2023rs5m}, and \cite{wang2024skyscript} have undertaken continual pre-training of CLIP on paired RS data, thereby extending its capabilities to encompass RS scene understanding. 

Concurrently, the advent of large language models (LLMs), exemplified by ChatGPT \citep{brown2020language, ouyang2022training}, alongside other models \citep{brown2020language, ouyang2022training, du2021glm, jiang2024mixtral}, has significantly enriched the landscape of NLP. These LLMs possess the remarkable ability to interact with humans through natural language, facilitate in-context learning, instruction following \citep{ouyang2022training}, and reasoning \citep{wei2022chain}, showcasing their versatility across diverse linguistic tasks. Motivated by the power of LLMs, researchers have proposed to align visual models with large language models to achieve the SOTA performance on image captioning \citep{li2022blip}, visual question answering (VQA) \citep{li2023blip, zhu2023minigpt, dai2024instructblip, liu2024visual}, visual grounding \citep{wang2023cogvlm}, and etc. In the RS domain, efforts have mirrored these trends. For instance, RSGPT \citep{hu2023rsgpt} fine-tuned InstructBLIP \citep{dai2024instructblip} on a self-built dataset to achieve state-of-the-art performance on RS image captioning and VQA tasks. Remote Sensing ChatGPT \citep{guo2024remote} utilizes ChatGPT to interpret user queries, plan remote sensing interpretation tasks, and generate final responses for users. Additionally, GeoChat \citep{kuckreja2023geochat} fine-tuned LLaVA-1.5 \citep{liu2024visual} using LoRA \citep{hu2022lora}, enabling its support for multi-task and multi-round conversion with users to interact with RS imagery.

Despite the progress in developing vision language models tailored for RS tasks, a significant gap remains between these models and those designed for general purposes. One of the main differences between the two paradigms lies in the scale of the models and their training data. Several studies have empirically observed that model performance is positively correlated with the model size and the data scale \citep{kaplan2020scaling, riquelme2021scaling, tay2021scale, zhai2022scaling, cherti2023reproducible}. Notably, the CLIP study trained its largest model, ViT-L/14, on a private WebImageText (WIT) dataset containing 400 million image-text pairs, resulting in a substantial leap in performance compared to models trained on smaller datasets. The success of CLIP has motivated the development of subsequent VLMs with increased model size and data scale for better performance. For instance, ALIGN \citep{jia2021scaling} trained a large EfficientNet-L2 image encoder on a private dataset of 1.8 billion image-text pairs. CoCa \citep{yu2022coca} used the ALIGN dataset as well as a private dataset called JFT-3B, containing 3 billion image-text pairs, to train a ViT-g/14 visual encoder and a language model. PaLI \citep{chen2022pali} further expanded this approach by training a multi-language, multi-task VLM using ViT-e (4B parameters) and mT5-XXL (13B parameters) on a private dataset of 29 billion image-text pairs.

The ever-growing model sizes and their data demands have quickly surpassed the capacity of earlier open image-text datasets like Microsoft COCO \citep{lin2014microsoft}, Visual Genome \citep{krishna2017visual}, YFCC100M \citep{thomee2016yfcc100m}, and CC3M \citep{sharma2018conceptual}, which are insufficient for researchers without access to large-scale private datasets. Recently, the availability of large-scale multimodal data via the Internet has led to the creation of even larger image-text datasets, such as LAION-400M \citep{schuhmann2021laion}, COYO-700M \citep{kakaobrain2022coyo-700m}, and LAION-5B \citep{schuhmann2022laionb}. These datasets have significantly enriched the VLM research community, enabling the development of many advanced VLMs \citep{li2023blip, wang2023cogvlm, zhu2023minigpt}. 

\begin{figure}[ht]
     \centering
     \includegraphics[width=0.48\textwidth]{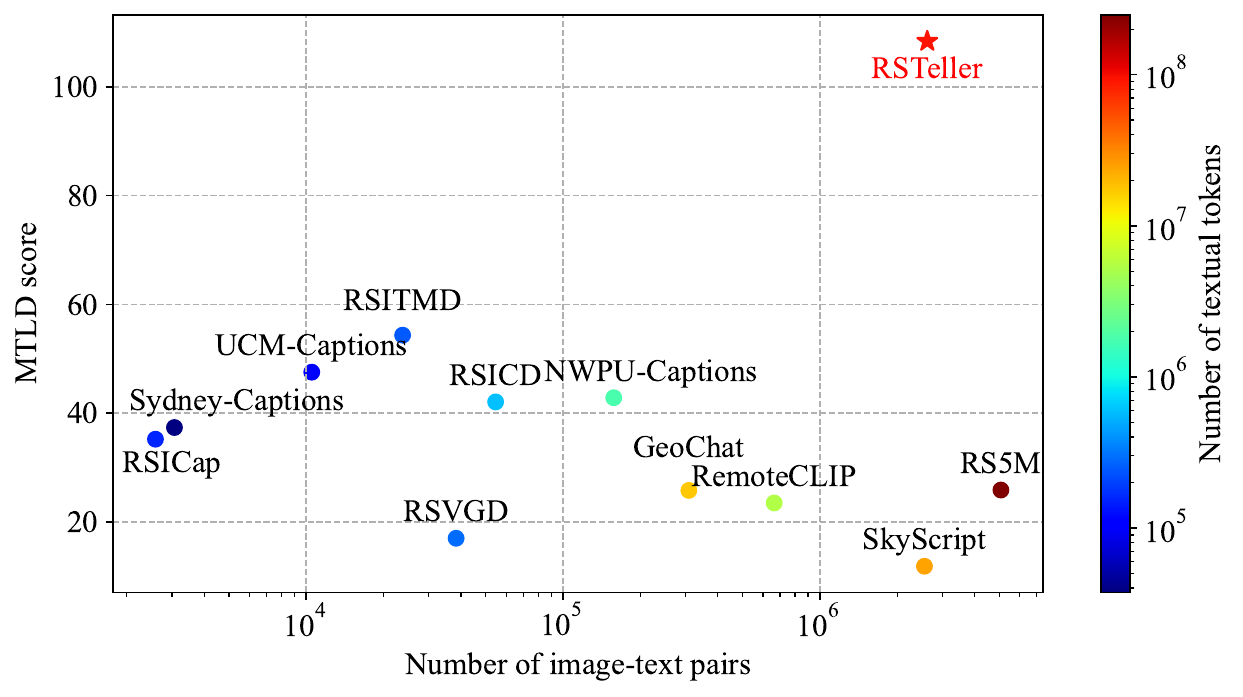}
     \caption{Comparison of existing image-text multimodal RS datasets. The Measure of Textual Lexical Diversity (MTLD) scores quantify the semantic richness of the captions. All captions were tokenized using the Natural Language Toolkit (NLTK) word tokenizer, and the color of each point indicates the scale of the tokens. The proposed RSTeller dataset is marked with a star. Its MTLD score is more than twice that of the next most diverse dataset, while also being substantial in data size.}
     \label{fig:dataset_comparison}
\end{figure}

The RS community has also made significant efforts to address the data scarcity problem for training VLMs for RS tasks. To facilitate readers' understanding of the semantic richness and scale of related works, we present a comparison in \cref{fig:dataset_comparison}. This figure illustrates the semantic richness of each dataset proposed in previous works, as quantified by the Measure of Textual Lexical Diversity (MTLD) score \citep{mccarthy2010mtld}, alongside the dataset scale, represented by both the number of textual tokens and the number of image-text pairs. Before the rise of large-scale pre-trained multimodal models like CLIP, several image-text datasets were developed, including UCM-Captions \citep{qu2016deep}, Sydney-Captions \citep{qu2016deep}, RSICD \citep{lu2017exploring}, RSITMD \citep{9437331}, NWPU-Captions \citep{9866055}, and RSVGD \citep{zhan2023rsvg}. These datasets predominantly feature images paired with concise captions that provide basic descriptions of primary scenes or objects. While the RSICap dataset \citep{hu2023rsgpt}, introduced in the RSGPT work, sought to deliver richer semantic descriptions, its relatively low MTLD score highlights a limited diversity in lexical usage and sentence structures. Moreover, all the aforementioned datasets are constrained by scale, with NWPU-Captions being the largest, comprising 31,500 images and 157,500 image-text pairs. The limited size of these datasets poses challenges for training VLMs like CLIP to perform well on RS tasks. 

To address the data scarcity problem, \cite{liu2024remoteclip} compiled a RemoteCLIP dataset from three different types of RS tasks (segmentation, detection, and retrieval) across 17 individual datasets. They used handcrafted rules to convert segmentation and detection annotations into multiple linguistic captions, resulting in a dataset with 165,745 images and 828,725 image-text pairs. Additionally, \cite{zhang2023rs5m} proposed leveraging existing large-scale image-text datasets by filtering data related to RS and captioning several RS classification datasets with a pre-trained image caption model, creating an RS-related multimodal dataset called RS5M with 5 million image-text pairs. Training VLMs on these larger RS datasets has demonstrated improvements in their performance. However, there are limitations in building datasets from existing ones, as the upper bound is restricted by the quantity of available well-crafted datasets, which is limited compared to the vast Internet resources. In contrast, datasets like LAION \citep{schuhmann2021laion, schuhmann2022laionb} were collected at an Internet scale, offering significantly larger amounts of data.

Despite the shortage of direct RS-related image-text datasets, there is an abundance of publicly available RS data. The Google Earth Engine (GEE) platform\footnote{https://developers.google.com/earth-engine} \citep{gorelick2017google} provides a rich source of RS imagery, with comprehensive coverage of the Earth's surface and access to a diverse range of aerial imagery from multiple sources. These sources vary in ground sampling distances, spectral bands, and sampling times, making them ideal for large-scale imagery data collection. On the other hand, OpenStreetMap (OSM)\footnote{https://wiki.openstreetmap.org} \citep{haklay2008openstreetmap} is a collaborative project that collects geo-elements worldwide, providing a rich source of geo-referenced data. OSM serves as a semantic source where geo-elements are annotated with descriptive tags, each composed of a key-value pair. The key describes a topic, category, or feature type (e.g., highway or name), while the value provides details for the specified feature (e.g., {\textbf{name}: Jeff Memorial Highway}). OSM currently contains over 96,000 distinct keys and over 155 million distinct tags, making it a valuable resource for harvesting geo-related linguistic data. Recently, \cite{wang2024skyscript} proposed using GEE and OSM as data sources to build a large-scale image-text dataset for RS called SkyScript, comprising 2.6 million image-text pairs. They selected valid OSM elements for given RS images and designed a heuristic caption-assembling method to convert tags into captions. The dataset exhibits great generality for downstream tasks when training a CLIP model and shows scalable potential since it uses openly available data. However, the captions in the dataset are all simple phrases rather than semantically rich sentences, as they are merely combinations of the tag words. This limitation affects the semantic richness of the dataset and its further applicability.

Using LLMs to annotate data has become increasingly common as the performance of these methods improves. For instance, \cite{li2023monkey} proposed using multiple VLMs and ChatGPT to refine the captions in image-text datasets, generating detailed image captions. To address the scarcity of well-annotated multimodal data, \cite{wang2023caption} introduced Caption Anything (CAT), which utilizes the Segment Anything Model (SAM) \citep{kirillov2023segment} and ChatGPT. CAT supports diverse visual and language controls to produce controllable captions. In another approach, \cite{fan2024improving} used LLaMa-7B \citep{touvron2023llama} to rewrite the captions in the training set of a CLIP model as a method of data augmentation, and validated the effectiveness of these rewrites in the downstream tasks. In the field of RS, GeoChat \citep{kuckreja2023geochat} employed Vicuna-1.5-7B \citep{chiang2023vicuna} to generate multi-round question and answer pairs for existing labeled RS datasets focused on object detection, visual question answering, or scene classification tasks. This resulted in a dataset with 306,000 image-instruction pairs, enabling the VLM to follow human instructions and handle multiple RS-related tasks. These methods demonstrate the feasibility and value of generating multimodal data with LLMs, inspiring us to design a large-scale multimodal dataset for RS using LLMs.

\subsection{Contributions}

In this work, we leverage openly available remote sensing data as an abundant resource for multimodal data generation. Specifically, we utilize the GEE platform to collect unlabeled aerial images, and the OSM platform as our semantic source. We employ two variant open-source LLMs to interpret plain OSM data. The OSM tag contents and their combinations are used to generate fluent, semantically rich descriptions for the aerial images. To ensure reproducibility and facilitate further scaling up of a dataset suitable for VLM training, we develop a comprehensive dataset generation workflow that encompasses raw data fetching, LLM-based captioning, and final dataset compilation.


Through this workflow, we introduce RSTeller, a multimodal dataset consisting of over 1.3 million remote-sensing images, each accompanied by two descriptive captions. The semantic richness of these captions, measured using the MTLD score \citep{mccarthy2010mtld}, is about twice that of existing datasets (\cref{fig:dataset_comparison}). Given the computational demands of LLM inference, our dataset provides a ready-to-use resource for researchers, eliminating the need for individual data generation. 

Extensive experiments were conducted to validate the proposed methods and provide valuable insights for researchers training VLMs on RS data. First, we demonstrated the effectiveness of RSTeller by the continual pre-training of multiple VLMs and evaluating their performance on downstream tasks. Second, we performed an ablation study to analyze the impact of data properties on model training. Third, we examined the relationship between VLM performance and the scale of RS-specific training data, underscoring the benefits of scaling data further. The proposed data generation workflow and dataset are expected to be instrumental in advancing more sophisticated VLMs within the remote sensing field.

\subsection{Paper Structure}
The structure of this paper is as follows: \cref{sec:workflow} details the proposed automatic data generation workflow. \cref{sec:dataset} describes and discusses the generated dataset, RSTeller, exploring its attributes in detail. \cref{sec:experiments} presents experiments evaluating the effectiveness of the proposed dataset, an ablation study on data properties, and an analysis of VLM performance with respect to data scale. \cref{sec:discussions} provides some findings and suggestions according to our experiments. \cref{sec:limitations} outlines the limitations of the proposed work. Finally, \cref{sec:conclusions} offers a summary of the research.

\section{Automated Workflow for Dataset Generation}
\label{sec:workflow}

\begin{figure*}[ht]
     \centering
     \includegraphics[width=2\columnwidth]{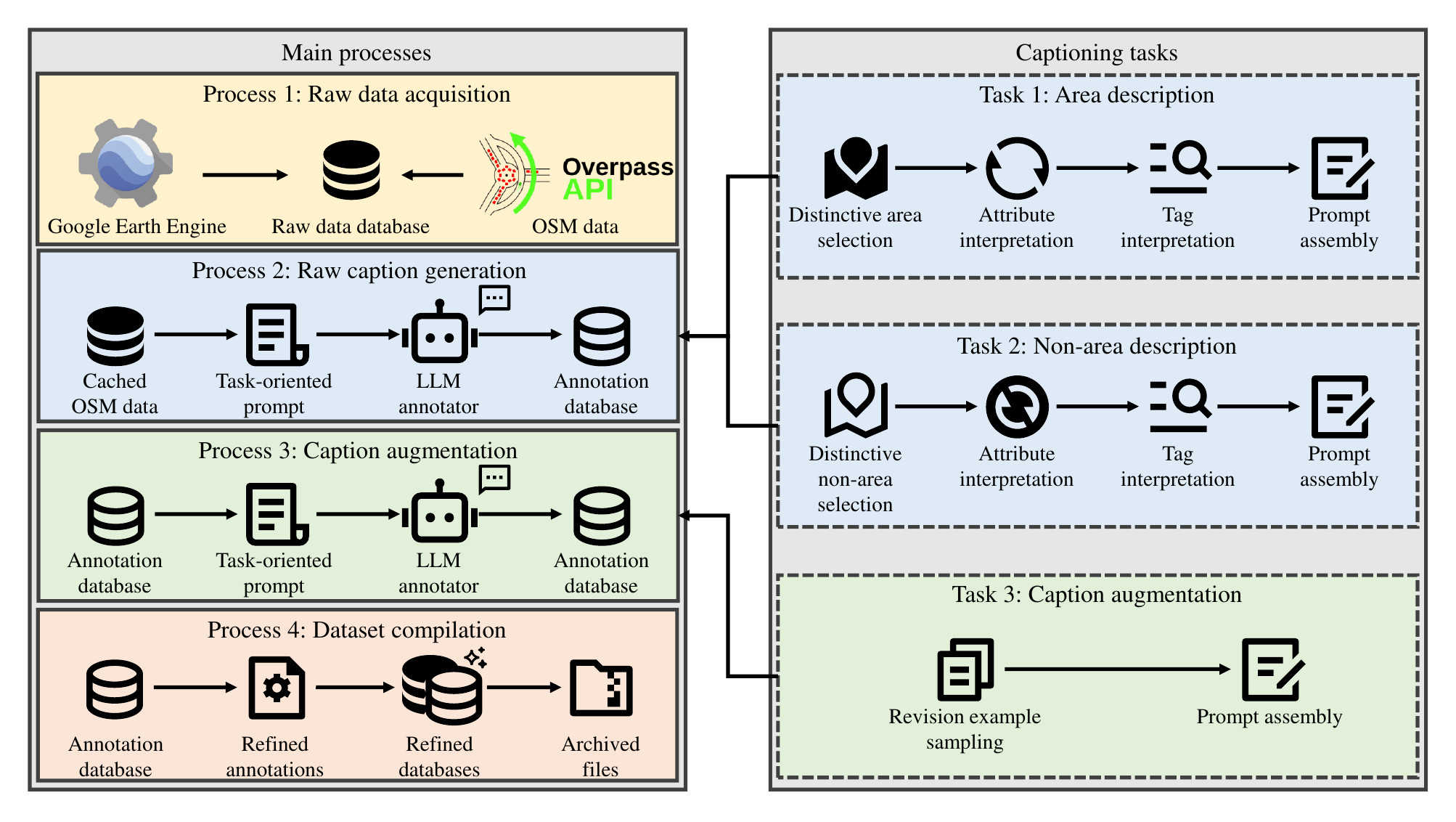}
     \caption{Workflow for dataset generation. The left section outlines the main processes: data fetching, LLM-based captioning, and final dataset compilation. The right section details the specific captioning tasks that enhance the quality and richness of the generated captions.}
     \label{fig: workflow}
\end{figure*}


In this section, we detail the automated workflow developed to generate a large-scale, RS-related multimodal dataset from openly available data. The overall workflow can be divided into two parts and is shown in \cref{fig: workflow}. The left panel of the figure illustrates the main processes, including data fetching, LLM-based captioning, and final dataset compilation. All the processes can be run in parallel for the maximum efficiency. The right panel provides details of the captioning tasks that facilitate the generation of semantically rich captions.

\subsection{Main Processes}

\subsubsection{Raw Data Acquisition}

The first process in the workflow is to fetch raw data from the GEE and OSM platforms into well-defined local databases. For ease of reference, we designate these databases as the ``raw data database,'' as depicted in Process 1 of \cref{fig: workflow}. This database is crucial for preserving the downloaded data and enabling asynchronous handling of different processes to accelerate the entire workflow. The GEE platform provides a vast collection of aerial images, each typically covering a wide range of geographical locations with tens or hundreds of thousands of pixels. The sizes of these images are too large for direct use in training a VLM, necessitating the cropping of images into smaller patches. GEE offers several flexible APIs for patch extraction and data download, from \texttt{ee.Image.getThumbURL} for JPEG or PNG formats to \texttt{ee.Image.getDownloadURL} for GeoTIFF or NumPy formats. For detailed API usage instructions, the official documentation\footnote{https://developers.google.com/earth-engine/apidocs} is recommended.

To enhance file system performance, a database such as MongoDB is preferred and used in this work for storing the downloaded image patches, as a plain file system may be inefficient for large data indexing and searching. In addition to the imagery data, metadata such as patch coordinates and image timestamps are also required and carefully preserved for OSM data downloading and further analysis. An SQL-based database like MySQL or PostgreSQL is recommended for storing these metadata.

The OSM data is another important source of semantic information for remote sensing. It can be queried via the Overpass API\footnote{https://wiki.openstreetmap.org/wiki/Overpass\_API} using specific OverpassQL queries. OSM contains three types of geo-elements: nodes, ways, and relations. As nodes are points and too small to be seen in an RS image, we focus only on ways and relations. A notable challenge in utilizing the Overpass API is that it processes ways and relations as collections of nodes. When using the bounding box filter, one of the most common approaches to querying spatial data provided by Overpass, only geometries with at least one node located within the bounding box are retrieved. As a result, large enclosing geometries, such as lakes or forests, that fully contain the bounding box but lack any nodes within it are not captured. To address this limitation, we implemented a two-step method to ensure comprehensive inclusion of relevant geo-elements. In the first step, we queried ways and relations using the bounding box filter to retrieve all geometries that satisfy the filter's policy. This step captures many essential geo-elements but may miss larger enclosing geometries, as described. In the second step, we performed an additional query using the \texttt{is\_in} operator, specifically targeting geometries tagged as ``area,'' to identify elements that enclose the center point of the image patch. This approach ensures the inclusion of large geo-elements that might have been overlooked in the first step, thereby providing a more complete representation of the image content.

The geographical contents of the image patches may evolve over time. Since the OSM data is not updated in real-time, we queried the OSM data for each image patch up to one month after it was captured by the sensor to best reflect the current image content. However, querying from the official endpoint\footnote{https://overpass-api.de/api/interpreter} requires following a strict usage policy, allowing fewer than 10,000 queries and 1 GB of data per day. Alternative third-party backends, such as the Kumi Systems Overpass Instance\footnote{https://overpass.kumi.systems/api/interpreter}, offer more lenient usage policies but may be affected by server overload and network traffic. Based on our experience, the Kumi endpoint enabled the retrieval of OSM data for approximately 100,000 patches daily, each covering an area of 72,253.44 square meters. Given the computational demands of captioning valid patches and the potential for incomplete OSM coverage, querying remote servers on-demand was deemed inefficient. Caching OSM data locally prior to captioning or deploying a local Overpass server can significantly reduce query times and enhance overall efficiency. In this study, we opted to cache data from the Kumi endpoint due to the technical challenges and resource requirements associated with setting up a local Overpass server. By caching sufficient data in advance, we ensured uninterrupted captioning without the need for additional infrastructure. For larger-scale projects with sufficient computational resources and expertise, deploying a local Overpass server remains a recommended approach to further optimize scalability and performance.

\subsubsection{Raw Caption Generation}

Once the data fetching process is underway and a sufficient amount of raw data has been acquired, the workflow proceeds to the generation of raw captions. This process uses LLMs, specifically Mixtral-Nemo-Instruct (12B) and Mixtral-Small-Instruct (22B) \citep{jiang2024mixtral}, to interpret and describe the OSM data for each image patch based on predefined tasks randomly. Model size was chosen according to the balance between caption quality and computational cost. The primary focus of our task is rephrasing raw materials into fluent sentences, with some minor reasoning to infer the surroundings and appearance of OSM elements. Since this reasoning is relatively simple compared to tasks like solving complex mathematical problems or multi-step logical reasoning, smaller models are sufficient for our needs. They provide high-quality outputs while minimizing computational costs, making them ideal for large-scale dataset generation. The Mixtral family was selected for its strong performance across a range of natural language tasks, accessibility, and flexible use policy. Additionally, using models of different sizes and versions introduces diversity in the captions, which could benefit future research on model variation and its impact on generated data.

In this process, a patch that has not yet been captioned is first sampled from the raw data database. The corresponding OSM data is then fetched from the cached OSM database. The OSM data undergoes processing based on our predefined tasks, which will be discussed in Subsection \ref{sec:CaptioningTasks}. This step checks whether the OSM data contains tags that can be used to generate captions for the patch. Patches that contain no relevant way or relation elements other than administrative borders, underground features, or where the OSM elements are too small to be visualized, are deemed unusable. These unusable patches are discarded and labeled as such in the raw data database. Only OSM data from valid patches are retained for further processing.

The predefined tasks further interpret the attributes of the OSM elements, forming task-oriented prompts for the LLM. The LLM takes these prompts as input and generates captions for the image patches. Finally, the generated captions are saved in an annotation database, and the status of the image patches is updated in the raw data database to reflect their completion.

\subsubsection{Caption Augmentation}




Data augmentation in the linguistic aspect has been shown to enhance the training of VLMs \citep{fan2024improving}. Unlike data augmentation in the visual domain, such as color jittering, noising, and random cropping, which can be easily implemented online during model training, augmenting the language aspect with consistent linguistic semantics and varied tones is resource-intensive, requiring the use of an LLM for high-quality results. Therefore, we incorporate a caption augmentation process into our workflow to generate one revision of each caption.

During the caption augmentation process, captions that have not yet been revised are sampled from the annotation database. A specific revision task is defined to generate a prompt for the LLM based on the original caption and its initial task. The details of these tasks will be discussed in Subsection \ref{sec:CaptioningTasks}. The same LLMs, Mixtral-Nemo-Instruct and Mixtral-Small-Instruct, are used for this process. It takes the prompt as input and generates a revised caption for the image patch. The revised caption is then saved in the annotation database, and the status of the image patch is updated in the raw data database to reflect its completion.

This caption augmentation process is repeated until the desired number of revised captions is generated for each image patch. In RSTeller, our proposed dataset under this workflow, each image patch has \replaced{two different captions}{at least two captions and up to five captions} in total.

\subsubsection{Dataset Compilation}

As the annotation database populates and scales to the desired size, the final process in the workflow involves refining the annotations and compiling a dataset suitable for VLM training at scale. 

During the inference of the LLM, certain errors in the generated captions are inevitable and can compromise the overall quality of the dataset. These errors are broadly categorized into two types: fixable errors and unfixable errors. Fixable errors typically follow recognizable patterns, such as repeated sentences appended after the caption or undesired outputs generated due to specific patterns in the input prompt. To address these, we implemented a systematic process. First, captions are segmented into individual sentences, and the uniqueness of each sentence is verified to detect and remove duplications. Second, common patterns of undesired outputs are identified through a comprehensive survey of the dataset. Based on these observations, we developed and applied a set of regular expressions to automatically detect and eliminate such errors, ensuring that the retained captions are concise and relevant. Unfixable errors, such as blank responses, invalid symbols, unexpected outputs, or duplicate captions generated for the same image patch, cannot be corrected and must be removed from the dataset. Similar to the handling of fixable errors, we analyzed the dataset to identify overarching patterns of unfixable errors and utilized handcrafted regular expressions to detect and eliminate them efficiently. The occurrence of these errors is primarily attributed to the limitations of the LLM, such as sensitivity to prompt design or variability in generative consistency. Moreover, error patterns may differ across various LLMs due to differences in their training data and architectures. However, the frequency of errors can be significantly reduced through several strategies: refining input prompts to provide clearer guidance or employing more advanced LLMs with enhanced capabilities. As LLM technology continues to advance, these measures are anticipated to further enhance the reliability and semantic accuracy of generated captions, thereby consistently elevating the quality of synthetic data produced by LLMs.

Once the captions are refined, they are used to update the annotation database for future use, making the data ready for final compilation. Training a typical deep learning model on a given dataset usually involves randomly shuffling the data and sequentially iterating through the shuffled dataset. However, storing image patches and their annotations individually on the local disk and reading them randomly during training is time-consuming, especially when the dataset comprises millions of images. Additionally, training a VLM typically requires large batch sizes with multiple GPUs, resulting in an I/O bottleneck.

To address this issue and facilitate the distribution of the dataset over the internet, we propose compiling the dataset into multiple shards of tar files following the WebDataset format\footnote{https://github.com/webdataset/webdataset}. Each tar file comprises a set of image patches and their annotations. The RSTeller dataset, proposed in this work, is provided in this format. During training, multiple shards of the dataset can be prefetched into memory, then randomly shuffled and iterated through. This approach effectively improves I/O performance and results in better training efficiency. 

\begin{figure*}[ht]
     \centering
     \includegraphics[width=2\columnwidth]{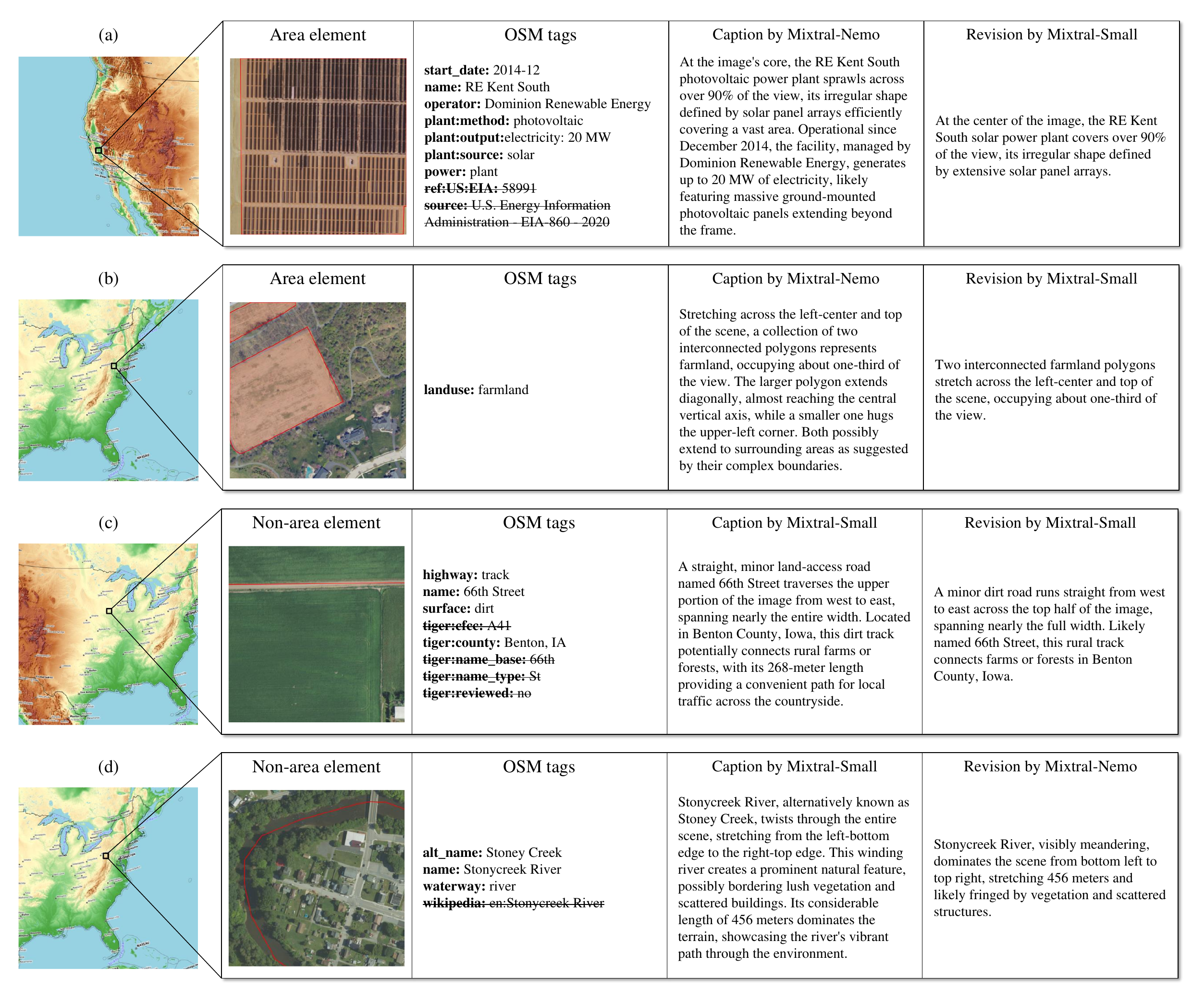}
     \caption{Inspection of the RSTeller dataset, showcasing four image patches with corresponding OSM tags, captions, and their revisions. The map highlights sample locations, while the figure illustrates OSM area and non-area elements with masked boundaries or lines. Crossed-out tags indicate filtered elements excluded from captioning. Captions represent results from Task 1 or Task 2, while revisions are outcomes of Task 3.}
     \label{fig:dataset_inspection}
\end{figure*}

\subsection{Captioning Tasks}
\label{sec:CaptioningTasks}
\subsubsection{Descriptive Tasks}

The objective of the descriptive task is to generate detailed and contextually accurate descriptions of the OSM data associated with each image patch. These descriptions should capture the essential features and elements present within the patch, such as geographical landmarks, infrastructure, vegetation, and other notable features. These descriptions serve as the raw captions and are produced in Process 2. 

In a single image patch, there usually are a number of OSM elements, each with multiple attributes and tags. The information provided by all the OSM elements in an image patch can be overwhelming, making it challenging for an LLM to generate a concise and accurate description. To streamline the task and reduce the potential for hallucination, we define a focused task that captions only one distinctive OSM element per image patch. By reducing the input context length and concentrating on a single element, we ensure that the LLM processes relevant information without being overwhelmed by extraneous details. This approach simplifies the generation process and enhances the semantic accuracy of the captions. 

In OSM, a two-dimensional object bounded by one or more linear ways and marked by appropriate tags is recognized as an ``area''. The simplest areas are defined by closed ways, having the same start and end node. More complex areas are defined using multipolygon relations, which may specify multiple boundary ways. Following this approach, we categorize all OSM elements in an image patch into two groups: ``area'' and ``non-area''. The ``area'' category includes OSM elements like land use, natural features, and buildings, while the ``non-area'' category includes elements not recognized as an ``area,'' such as roads, coastlines, and other linear features. Given the significant differences in geometry and attributes between these two categories, we have designed two distinct tasks: Task 1 for the ``area'' category and Task 2 for the ``non-area'' category. \cref{fig:dataset_inspection} presents examples from the RSTeller dataset, showcasing image patches, selected OSM elements, and corresponding captions. The selected OSM elements are highlighted in red on the image patches. Image patches (a) and (b) are captioned based on the ``area'' category, while the others are captioned based on the ``non-area'' category.

\textbf{Task 1 (Area description):} As is shown in \cref{fig: workflow}, this task follows four steps: distinctive area selection, attribute interpretation, tag interpretation, and prompt assembly. 

\textit{Distinctive area selection}: In this step, OSM elements smaller than 0.05 of the image patch area are filtered out to exclude insignificant features. From the remaining elements, if more than three candidates exist, one is randomly selected from the top three by area; otherwise, selection is made from all available elements. This approach highlights spatially dominant features while enhancing diversity and ensuring broader geo-element representation across image patches.

\textit{Attribute interpretation}: To enable robust and context-aware training for VLMs, we carefully selected attributes that encapsulate the critical geometric and semantic properties of geo-elements. These attributes ensure the dataset effectively captures essential spatial, morphological, and contextual details, enhancing the descriptive and inferential capabilities of VLMs. By striking a balance between granularity and interpretability, they optimize the dataset's utility across diverse remote sensing tasks. As summarized in Table \ref{tab: task 1 attributes}, we define five key attributes: coarse location, shape, normalized size, simplified geometry, and is-cropped. The coarse location attribute identifies the center of the element within a \(3 \times 3\) grid, assigning a label to each grid cell (e.g., ``left-top,'' ``center,'' ``right-bottom''). This attribute provides spatial context, helping VLMs map object positions relative to the image patch and supporting tasks requiring spatial reasoning. The shape attribute describes the approximate shape of the element based on its geometry, defined by four categories: ``square,'' ``rectangular,'' ``circular,'' and ``irregular''. This simplifies the representation of complex geometries, enabling models to recognize and differentiate structural patterns efficiently. These categories are determined by hard thresholding on the element's area-to-perimeter ratio and shape-to-area ratio. The normalized size attribute quantifies the relative scale of the element within the image patch, offering essential context for understanding the prominence and spatial importance of the element in the scene. The simplified geometry attribute encodes the spatial outline of the element in a compact, normalized format, preserving essential geometric characteristics while minimizing complexity. Although not directly included in the final captions, this attribute functions as supplementary input for the captioning LLM, enhancing its interpretation of the size, shape, and location attributes to produce more informative and contextually accurate descriptions. To avoid overwhelming the LLM annotators with extremely complex geometries, we use the Douglas-Peucker algorithm \citep{ramer1972iterative} to simplify the element's geometry. As some complex elements may be represented by multiple polygons, we encapsulate the coordinate pairs of each polygon within a list, and all polygons of an element are enclosed within braces. In this work, (0, 0) represents the bottom-left corner and (1, 1) the top-right corner, with all coordinates normalized to the range of [0, 1]. The is-cropped attribute indicates whether parts of the element extend beyond the image patch. This provides critical information about spatial boundaries, enhancing the model's ability to handle cropped or incomplete objects—a common scenario in remote sensing datasets. This attribute is determined by checking if the element visible in the image patch is cropped from a larger geometry.

In some cases, an area element can be represented as a multipolygon. This typically occurs due to complex geometries or when a larger geometry is cropped into separate parts. \cref{fig:dataset_inspection}(b) illustrates an example where a farmland area is split into two separate polygons. In such cases, we provide a coarse location interpretation for each polygon but focus on describing the shape of the largest one. The normalized size is calculated based on the total area of the entire multipolygon. Attributes such as simplified geometry and is-cropped are also derived from the overall multipolygon structure.

\begin{table}[htbp]
     \centering
     \caption{Attributes for distinctive area description in captioning Task 1}
     \resizebox{0.48\textwidth}{!}{
          \begin{tabular}{ll}
               \toprule
               Attribute & Example \\
               \midrule
               Coarse Location & ``left-top'', ``center'', etc. \\
               Shape & ``square'', ``rectangular'', ``circular'', ``irregular'' \\
               Normalized Size & ``0.300'', ``0.453'', etc. \\
               Simplified Geometry & ``\{[(0.000, 0.500), (0.120, 0.400), ...]\}'', etc. \\
               Is-cropped & True, False \\
               \bottomrule
               \end{tabular}%
     }
     \label{tab: task 1 attributes}%
\end{table}%

\textit{Tag interpretation}: OSM tags serve as the primary semantic source for the captioning task but can also introduce significant noise, which may adversely affect the performance of captioning models and lead to hallucinations. This noise primarily arises from two sources: irrelevant tags and ambiguous tags.

Irrelevant tags, such as ``TIGER," ``website," and other metadata fields, often contain ID digits or information unrelated to the visual features of the image. To mitigate this issue, we apply a rigorous tag filtering process using predefined regular expressions (see \ref{app:regex} for details). This filtering ensures that only meaningful and relevant data is passed to the LLM for caption generation, thereby minimizing the risk of hallucinations. The effectiveness of this filtering is illustrated in \cref{fig:dataset_inspection}, where unused tags are visually marked with strikethroughs.

\added{Ambiguity in OSM tags further complicates their direct use in captioning, as some tags can be vague and difficult for both LLMs and human annotators to interpret. For instance, the tag ``\textbf{man\_made}: works'' might refer to various man-made structures, but according to the OSM Wiki, it specifically denotes a factory or industrial production plant. To resolve such ambiguities and enhance the semantic clarity of the tags, we query the OSM Wiki for all tag descriptions corresponding to the selected elements. These descriptions are reformatted into a uniform and linguistically consistent structure, replacing the raw key-value pairs as input to the LLM. For tags with unbounded values, such as ``\textbf{name}: Jeff Memorial Highway,'' we focus on interpreting the tag key (e.g., ``name") rather than the specific value. This standardized and enriched representation ensures a comprehensive and accurate description of each selected element. An example of these two types of tag interpretation is provided in Table~\ref{tab: tag interpretation example}.}

\deleted{Ambiguity in OSM tags further complicates their direct use in captioning. For instance, the tag ``\textbf{man\_made}: works'' can be vague and difficult for both LLMs and human annotators to interpret, as it might refer to a variety of man-made structures. To address this, we leverage the OSM tag Wiki database, which provides detailed descriptions of tag meanings. According to the Wiki, ``\textbf{man\_made}: works'' specifically refers to a factory or industrial production plant. By incorporating these explanations, we resolve ambiguities and enhance the semantic clarity of the tags.}

\deleted{To standardize and enrich the input for the LLM, we query the OSM Wiki for all tag descriptions corresponding to the selected elements. These descriptions are reformatted into a uniform and linguistically consistent structure, replacing the raw key-value pairs. For tags with unbounded values, such as ``\textbf{name}: Jeff Memorial Highway,'' we focus on interpreting the tag key (e.g., ``name") instead of the value. This approach ensures a comprehensive and accurate description of each selected element. An example of these two types of tag interpretation is provided in Table \ref{tab: tag interpretation example}.}

\begin{table}[htbp]
     \centering
     \caption{Examples of tag interpretation}
     \resizebox{0.48\textwidth}{!}{
          \begin{tabular}{p{0.5 \columnwidth}p{0.6 \columnwidth}}
          \toprule
          Tag   & Interpretation \\
          \midrule
          \textbf{man\_made}: works & man\_made: works. The tag belongs to the tag group ``NULL''. This tag means: ``A factory or industrial production plant''. \\
          \textbf{name}: Jeff Memorial Highway & Its key is ``name'', which means ``the primary name: in general, the most prominent signposted name or the most common name in the local language(s).''. The tag belongs to a tag group ``names''. The tag value is Jeff Memorial Highway. \\
          \bottomrule
          \end{tabular}%
     }
     \label{tab: tag interpretation example}%
\end{table}%

\begin{figure}[ht]
     \centering
     \includegraphics[width=1\columnwidth]{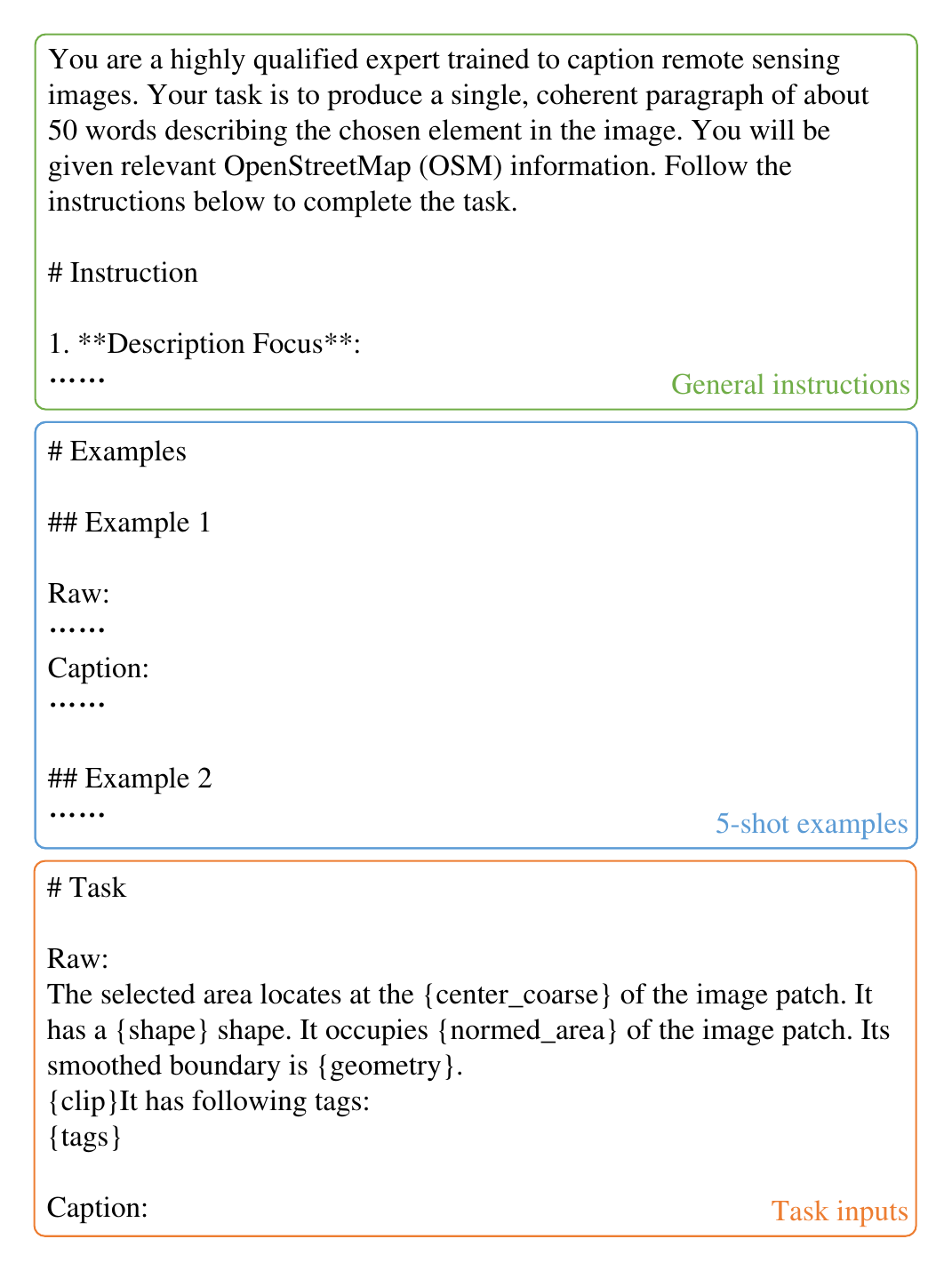}
     \caption{A brief showcase of the prompt template for Task 1. Details are replaced by ellipses for brevity. Placeholders are wrapped in braces and are to be filled dynamically by the task.}
     \label{fig: task 1 prompt template}
\end{figure}

\textit{Prompt assembly}: Once the attributes and tags have been interpreted, the final step in Task 1 is to assemble all the interpretations into a coherent prompt for the LLM. This involves organizing the interpreted attributes and tag descriptions into a structured format that the LLM can effectively process. In this work, we employ a template-based approach for prompt assembly, where we define a task-specific template to structure the prompt and fill in the placeholders with the interpreted attributes and tags. A brief showcase of the prompt template for Task 1 is presented in \cref{fig: task 1 prompt template}. The overall structure of the prompt is divided into three parts. First, general instructions are provided to the LLM, offering a high-level overview of the task followed by detailed guidance on completing it. Specifically, the instructions emphasize producing a fluent and natural description of the selected element. The LLM is instructed to describe the element based on its visual attributes, such as position, shape, approximate size, and notable features, while incorporating reasonable inferences about the surrounding context using cautious language (e.g., ``possible,'' ``likely''). This approach expands the task scope beyond the selected element to include the broader scene. Next, a five-shot example is included to illustrate the format and style of the desired output, helping the LLM understand the task requirements more clearly. The example follows the same format as the real task, consisting of a ``Raw'' and ``Caption'' pair. The ``Raw'' part includes attributes and tags interpreted from an example image patch, while the ``Caption'' part is the gold-standard caption from a human annotator and revised by GPT-4. The human annotator ensures that the caption captures domain-specific details accurately, but minor issues such as spelling, phrasing, or grammar errors can still occur. GPT-4 is then used to refine the caption, improving its fluency, consistency, and overall quality while maintaining its original meaning. Finally, the revised caption is double-checked by the annotator to confirm its correctness and domain alignment. This combined approach leverages both human expertise and GPT-4's linguistic capabilities to produce high-quality captions. Lastly, task-specific inputs are provided. The ``Raw'' part is a predefined context with multiple placeholders, where the interpreted attributes and tags are directly filled, except for the ``is-cropped'' attribute. If the ``is-cropped'' attribute is \texttt{True}, the placeholder is replaced with the sentence ``Some parts of the geometry extend beyond this ROI,'' otherwise, it is left blank. The ``Caption'' part remains blank, awaiting the LLM's continual writing. This assembly step ensures that the LLM's output is both precise and aligned with the intended use of the remote sensing data, facilitating effective and insightful captioning. The complete prompts for this task, as well as others used in this work, are publicly available in our GitHub repository for transparency and reproducibility.

\textbf{Task 2 (Non-area description):} Similar to Task 1, this task involves four steps. The element selection for Task 2 follows a similar structure to Task 1 but with different criteria. We first filter out elements whose length is smaller than 30\% of the square root of the image patch area. From the remaining elements, we randomly select one from the top three longest.


In the attribute interpretation step, we extract different attributes compared to Task 1 due to the geometrical differences between the ``area'' and ``non-area'' elements. The attributes for Task 2 are shown in Table \ref{tab: task 2 attribute}. Endpoint locations replace the coarse location for non-area elements, capturing positional information specific to linear features. This ensures accurate spatial descriptions for elongated structures like roads or rivers. The nine-grid system, the same as used in Task 1, assigns labels to the endpoints. The sinuosity attribute captures the geometric complexity of linear elements, offering insights into their curvature and form, which are crucial for distinguishing between infrastructure types like roads or waterways. Its categories including ``straight,'' ``curved,'' ``twisted,'' ``closed,'' and ``broken,'' determined by thresholding the ratio between the element's length and the distance between its endpoints. The normalized length attribute represents the relative length of the element compared to the square root of the image patch area, enabling models to assess the spatial extent of linear features relative to the image context. The length attribute is the actual length of the element in meters, offering precise quantitative information that enhances spatial and contextual reasoning. The orientation attribute, determined by calculating the angle between the endpoints and the horizontal axis, describes the direction of the element with categories like ``west-east,'' ``south-north,'' ``southwest-northeast,'' and ``northwest-southeast'', aiding the model in understanding spatial relationships and movement patterns within the image patch. For non-area elements exhibiting extreme curvature with a sinuosity value exceeding 1.5 (the defined threshold for distinguishing curved elements), the orientation is labeled as ``too curved or twisted to determine accurately.'' This ensures that the attribute remains meaningful and robust, even in cases of complex geometries. The simplified geometry attribute represents the simplified and normalized geometry of the element, following the same scheme as in Task 1. Finally, the is-cropped attribute indicates whether the element is cropped by the image patch. Given the differences in the nature of ``area'' and ``non-area'' elements, as well as the distinct attributes and their interpretation processes, we use separate icons to emphasize these variations in \cref{fig: workflow}. This differentiation enhances clarity and facilitates a better understanding of the respective workflows.

In some cases, a non-area element may also consist of separated lines, similar to the behavior in area elements. In such instances, we interpret the endpoint locations, sinuosity, and orientation based solely on the longest line within the element, while other attributes are derived from the entire element as a whole.

\begin{table}[htbp]
     \centering
     \caption{Attributes for distinctive non-area description in captioning Task 2}
     \resizebox{0.48\textwidth}{!}{
          \begin{tabular}{lp{0.85 \columnwidth}}
          \toprule
          Attribute & Example \\
          \midrule
          Endpoint Locations & ``(bottom-center, right-center)'', ``(left-bottom, right-bottom)'', etc. \\
          Sinuosity & ``straight'', ``curved'', ``twisted'', ``closed'', ``broken'' \\
          Normalized Length & ``0.351'', ``1.260'', etc. \\
          Length & ``94'', ``339'', etc. \\
          Orientation & `west-east'', ``south-north'', etc. \\
          Simplified Geometry & ``[{(0.211, 0.000), (0.320, 0.794), ...}]'', etc. \\
          Is-cropped & True, False \\
          \bottomrule
          \end{tabular}%
     }
     \label{tab: task 2 attribute}%
\end{table}%

The subsequent steps, tag interpretation and prompt assembly are largely similar between Task 1 and Task 2, with modifications in the prompt template to support the attributes specific to Task 2. To avoid redundancy, we do not elaborate on the details of these steps as they are consistent with those explained in Task 1. This structured approach ensures that each task is tailored to the unique characteristics of the elements being described, enhancing the accuracy and relevance of the LLM-generated captions for remote sensing data. The careful design of tasks and attributes enables a more effective analysis and utilization of the OSM data within the image patches.

\subsubsection{Caption Augmentation Task}


The caption augmentation task aims to enhance the diversity and quality of the initial captions generated for the image patches, creating a revision from the raw caption with different tones. This task, denoted as Task 3 in this work and illustrated in \cref{fig: workflow}, is relatively simple compared to previous tasks involving OSM element selection and interpretation, as it is purely a linguistic task and does not require specialized knowledge. Consequently, it comprises only two steps: revision example sampling and prompt assembly. 

Unlike previous tasks that utilized a fixed example for each task, this task necessitates a larger number of examples to produce diverse and informative captions. To this end, we constructed a meta-example set containing multiple raw captions and their corresponding revisions. In this work, we separately selected five raw captions from Task 1 and Task 2 and generated five revisions for each raw caption, altering the tone, inflection, word choice, and length of the original caption. These revisions were written by human annotators and polished by GPT-4 following the same approach as in Task 1 and Task 2. In Task 3, examples are dynamically sampled from the meta-example set, selecting all five raw captions of the corresponding task and randomly choosing one of the five revisions for each raw caption. The five raw and revision pairs are then shuffled and used as the five-shot examples for the prompt assembly step. This approach ensures that the revisions generated by the LLM are diverse and informative.

\begin{figure}[ht]
     \centering
     \includegraphics[width=1\columnwidth]{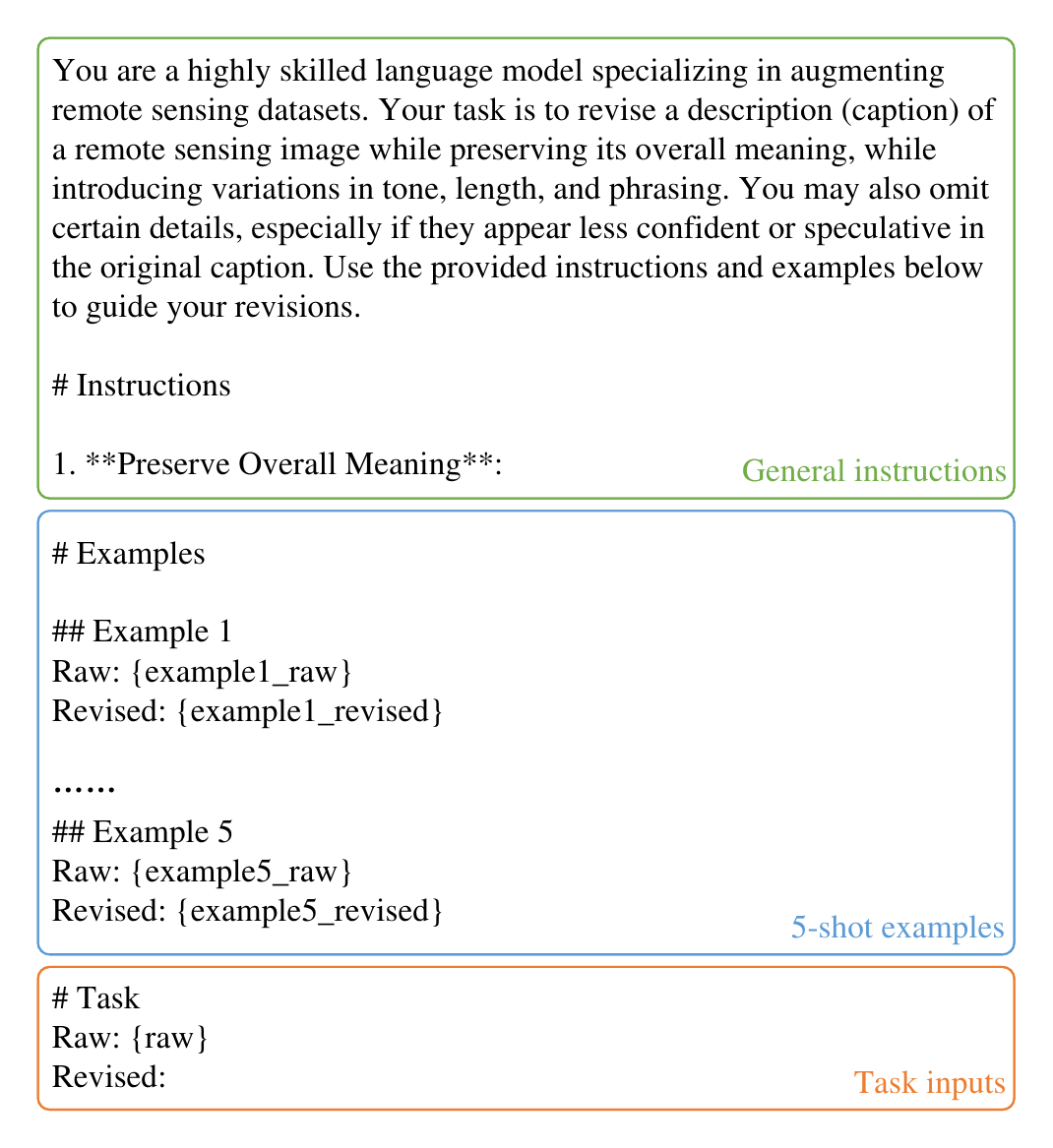}
     \caption{A brief showcase of the prompt template for Task 3. Details are replaced by ellipses for brevity. Placeholders are wrapped in braces and are to be filled dynamically by the task.}
     \label{fig: task 3 prompt template}
\end{figure}

In the prompt assembly step, a prompt template is utilized, as shown in \cref{fig: task 3 prompt template}. The overall structure of the prompt template is similar to that of Tasks 1 and 2, divided into three parts: general instructions, few-shot examples, and task inputs. The general instructions are crafted to guide the LLM in revising existing captions, emphasizing variations in tone, phrasing, and length while maintaining the core meaning. These instructions also provide clarity on the coordinate system used and highlight the role of examples in facilitating the task. The five-shot examples in the prompt template are defined by several placeholders, filled by sampled examples from the previous step. The task inputs include a placeholder for the caption to be revised, leaving the revised result empty for the LLM to fill in. 

This method ensures that the final augmented captions are of high quality, diverse, and aligned with the intended use of the remote sensing data, thus facilitating the effective and insightful analysis.

\section{RSTeller Dataset}
\label{sec:dataset}

Utilizing the proposed dataset generation workflow, we have developed the RSTeller dataset. The data acquisition and processing were conducted using a local server equipped with 128 cores, 512 GB of memory, and four NVIDIA A800 GPUs. To generate captions efficiently, we utilized two instances of Mixtral-Nemo-Instruct and two instances of Mixtral-Small-In\-struct, each assigned to a single GPU. The captioning tasks were dynamically distributed based on GPU occupancy to maximize parallel processing efficiency. To accelerate the captioning process, the vLLM framework \citep{kwon2023efficient} was utilized to host the large language models. The data acquisition process took approximately four months, during which we collected roughly 15 million image patches. Of these, approximately 7 million had corresponding OSM data fetched from the Kumi endpoint. The captioning process ran concurrently with data acquisition and processing, achieving an average throughput of about 200 thousand captions per day. This process spanned over two weeks and generated approximately 3 million captions. Ultimately, 2.6 million well-crafted image-text pairs were selected and compiled into the RSTeller dataset. The following sections provide detailed insights into the dataset.

\subsection{Dataset Overview}
\label{sec:dataset_overview}

RSTeller dataset comprises approximately 1.3 million remote-sensing image patches, each accompanied by two captions, resulting in roughly 2.6 million image-text pairs. The images are sourced from the National Agriculture Imagery Program (NAIP) on the GEE platform, with a ground sample distance (GSD) of 0.6 meters. Geographically, the dataset focuses on the United States, with images acquired via aircraft. Detailed attributes of the dataset are presented in Table \cref{tab:dataset_attributes}. 

\begin{table}[htbp]
     \centering
     \caption{Attributes of the RSTeller dataset generated by the proposed workflow}
     \resizebox{0.482\textwidth}{!}{
       \begin{tabular}{ll}
       \toprule
       \textbf{Attribute} & \textbf{Value} \\
       \midrule
       Number of Images & 1,309,926 \\
       Number of Image-Text Pairs & 2,619,852 \\
       Image Source & National Agriculture Imagery Program (NAIP) \\
       Ground Sample Distance (GSD) & 0.6 meters \\
       Pixel Dimensions & $448 \times 448$ pixels \\
       Image Bands & Red (R), Green (G), Blue (B) \\
       Image Capture Date Range & August 1, 2021 - November 26, 2022 \\
       Geographic Coverage & United States \\
       Image Acquisition Platform & Aircraft \\
       \bottomrule
       \end{tabular}%
       }
     \label{tab:dataset_attributes}%
\end{table}%

\begin{figure}[htbp]
     \centering
     \includegraphics[width=0.75\columnwidth]{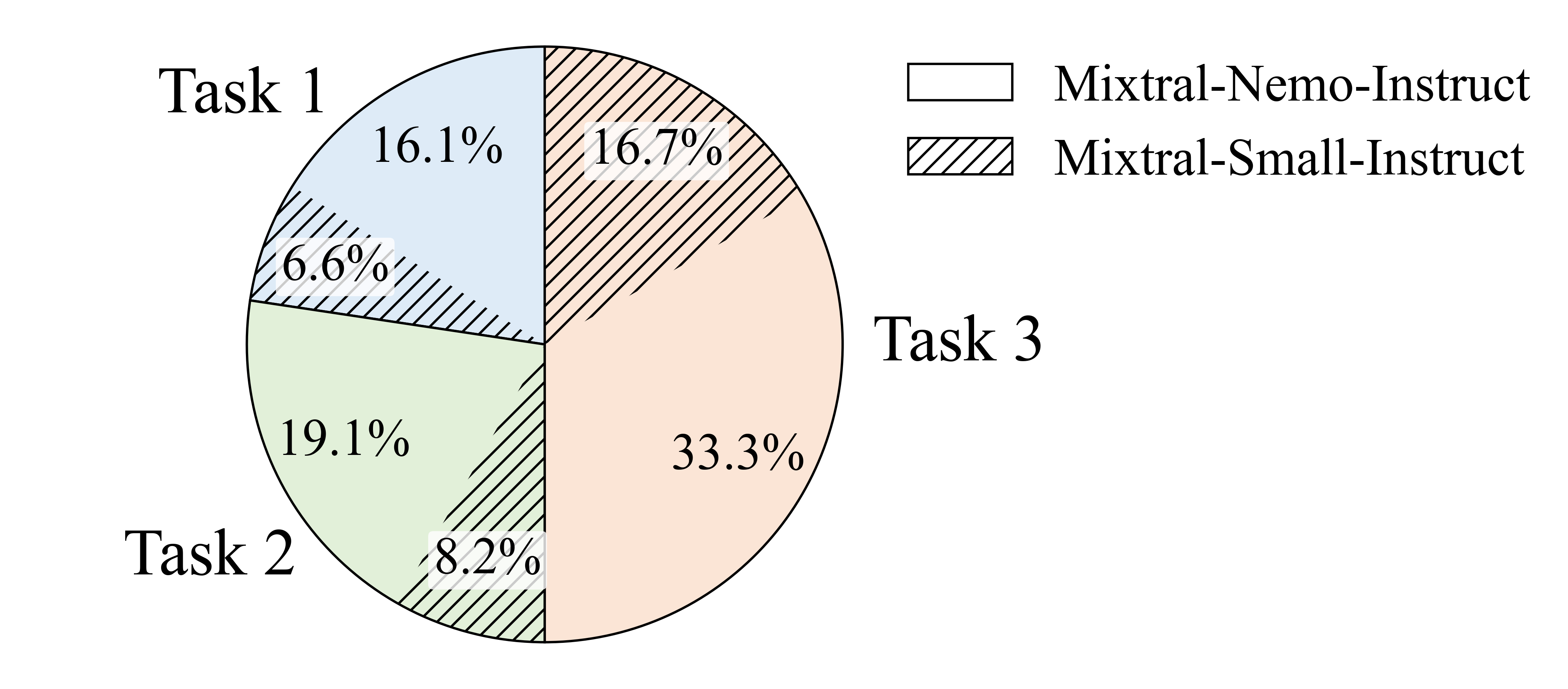}
     \caption{Distribution of captions across tasks in the RSTeller dataset. Solid colors indicate annotations generated by Mixtral-Nemo-Instruct, while shaded areas represent annotations from Mixtral-Small-Instruct.}
     \label{fig: caption_distribution}
\end{figure}

\cref{fig: caption_distribution} illustrates the distribution of captions across tasks in the RSTeller dataset. Task 1 captions (area elements) are slightly fewer than Task 2 captions (non-area elements). Task 3 captions, comprising revised versions of the raw captions, equal the combined total of Task 1 and Task 2 captions. Shaded regions indicate annotations generated by Mixtral-Small-Instruct, while solid regions correspond to annotations from Mixtral-Nemo-Instruct. The lower output of Mixtral-Small-Instruct is attributed to its larger model size and reduced overall throughput.


\begin{figure}[htb]
     \centering
     \includegraphics[width=1\columnwidth]{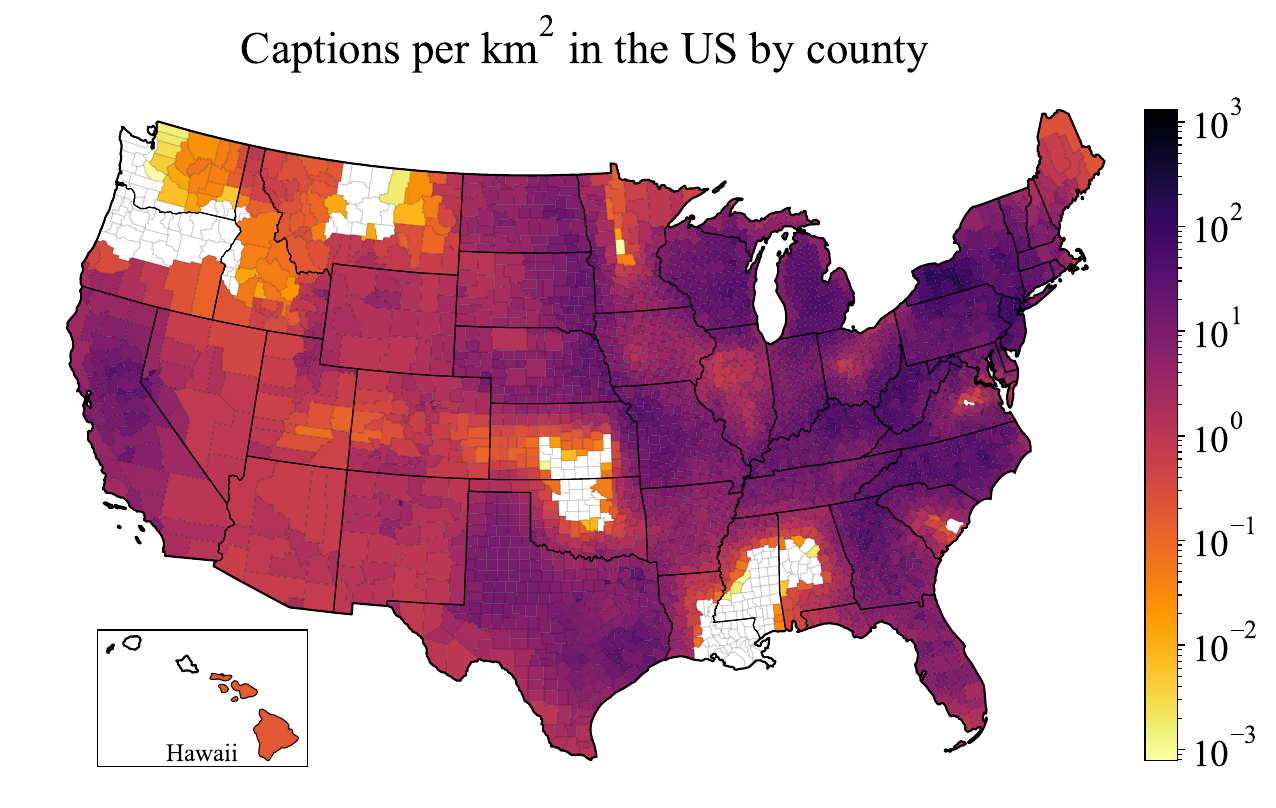}
     \caption{Geographical distribution of caption density across the United States.}
     \label{fig: caption_density}
\end{figure}

Rooted in NAIP, the dataset's images cover the majority of the continental United States and parts of Hawaii. We surveyed the dataset and plotted the caption density by county, as shown in \cref{fig: caption_density}, to provide a better understanding of the dataset's coverage and geographical distribution. Although the dataset covers most of the United States, some regions, such as Southern Mississippi and Northern Oregon, exhibit low or zero caption density. This is because the dataset was generated by sampling NAIP images based on their capture dates. For some regions, few NAIP images are available within the dataset's date range (August 1, 2021, to November 26, 2022), resulting in low caption density.

To facilitate qualitative analysis, we have provided a sample inspection of the dataset in \cref{fig:dataset_inspection}. This figure presents four image patches along with their corresponding OSM information and captions, sampled from various locations across the United States. Topographic base maps from OpenTopoMap\footnote{https://opentopomap.org} are included to provide a visual reference for the geographical locations of the samples. The figure then displays the image patches, selected OSM elements, their tags, and the captions under specific captioning tasks. Selected OSM elements are highlighted in red: area elements are highlighted with masks and enclosed by their respective boundaries, while non-area elements are highlighted with simple lines. OSM tags not incorporated into the captioning process are crossed out.

Overall, the RSTeller dataset is geographically and temporally diverse, covering a wide range of locations and time periods. It provides rich and varied semantics, making it suitable for training various VLMs for applications such as open-vocabulary classification, cross-modal retrieval, image captioning, and text-to-image generation in the remote sensing domain.

\subsection{Caption Analysis}


This section evaluates the captions generated by the LLM to identify key attributes and potential biases within the RSTeller dataset. Our aim is to provide valuable insights that will inform the effective utilization of the dataset in future applications.

\begin{figure}[ht]
     \centering
     \subfloat[\centering]{
          \includegraphics[width=0.48\columnwidth]{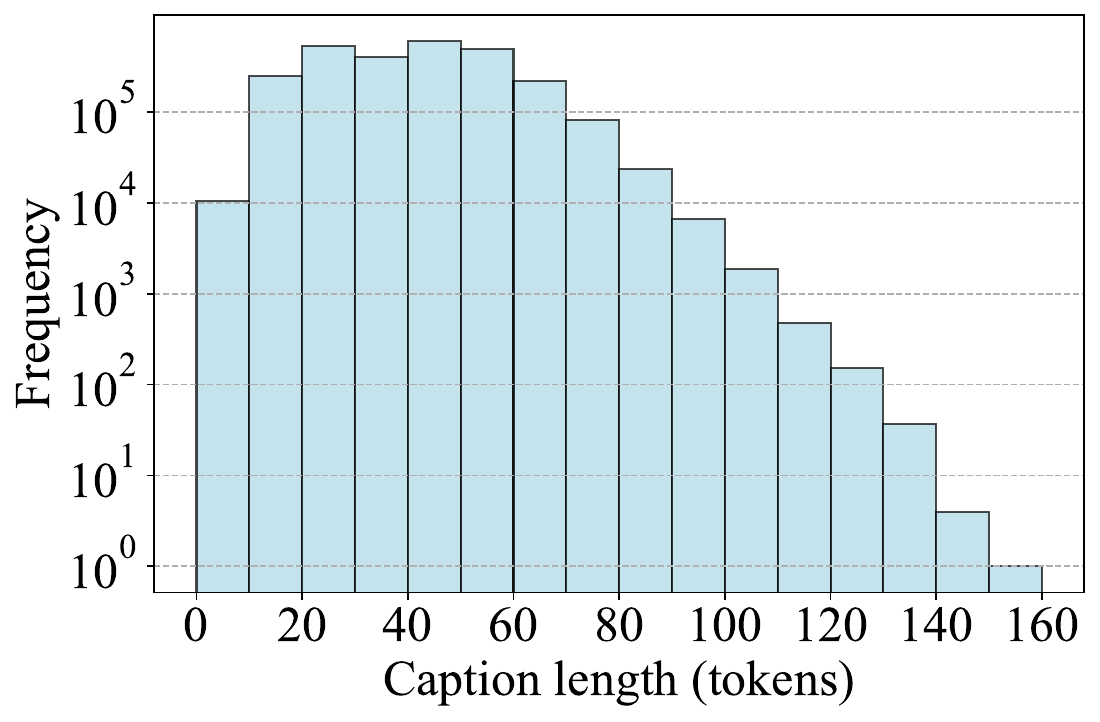}
     }
     \hfill
     \subfloat[\centering]{
          \includegraphics[width=0.48\columnwidth]{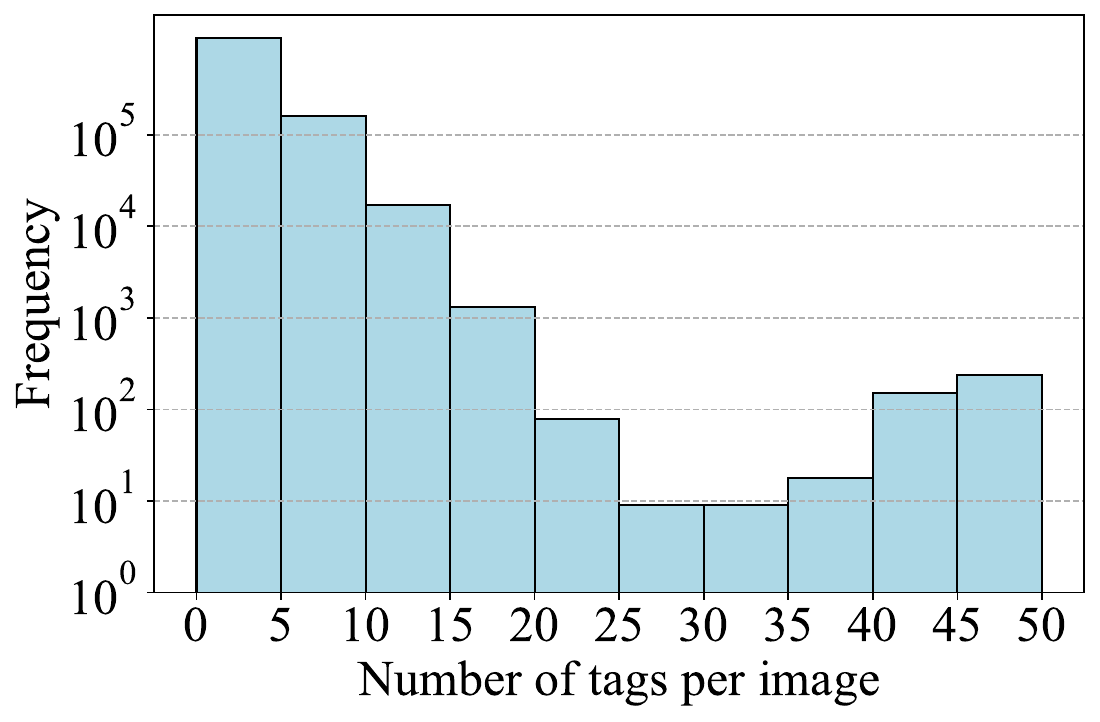}
     }
     \caption{Caption analysis of the RSTeller dataset. (a) Histogram of caption lengths. (b) Histogram of tag counts per image.}
     \label{fig: caption_analysis}
\end{figure}

First, we examine the distribution of caption lengths, determined by counting the number of tokens using the Natural Language Toolkit (NLTK) word tokenizer. The caption lengths vary from 2 to 155 tokens, with a median length of 42 tokens and an average length of 40.9 tokens and its distribution is depicted in \cref{fig: caption_analysis}(a). This variation underscores the diversity within the captions, which could enhance the robustness of VLM training. However, given that the CLIP model has a maximum input length of 77 tokens, captions exceeding this limit may compromise performance, highlighting the need for meticulous preprocessing.


We further evaluate metadata richness by analyzing the number of tags associated with each OSM element used for captioning. As shown in \cref{fig: caption_analysis}(b), the distribution displays a bimodal pattern, with a prominent peak in the 0-10 tags range, a secondary peak around 45-50 tags, and a noticeable dip between 25 and 35 tags. The secondary peak primarily arises from OSM elements related to protected areas, which often include tag groups starting with prefixes such as "nconemap" or "NCOS." These tag groups, encompassing attributes like land ownership, conservation management, and access restrictions, typically contain extensive metadata relevant to image captioning. As such, they are retained during preprocessing to preserve valuable information.

\begin{figure}[htbp]
     \centering
     \includegraphics[width=1\columnwidth]{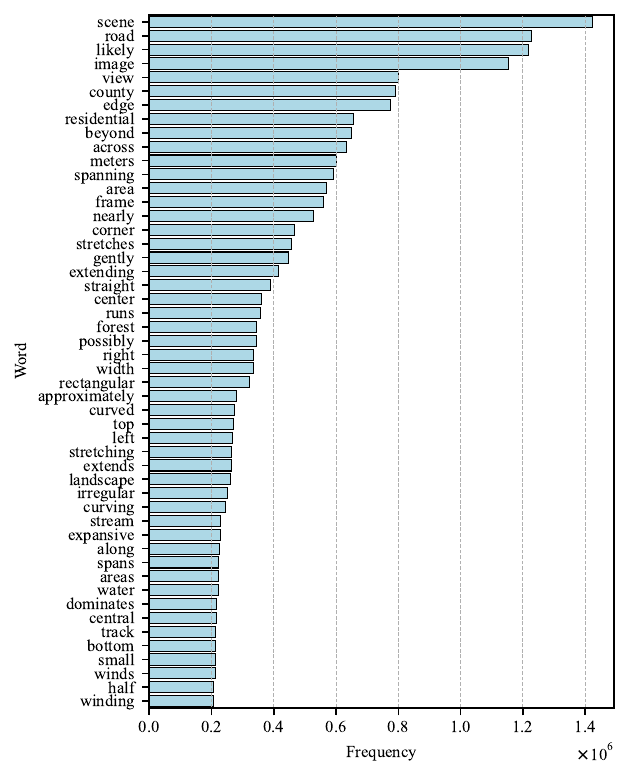}
     \caption{Top 50 most frequent words in the captions of the RSTeller dataset.}
     \label{fig: top_word_freq}
\end{figure}

Subsequently, we scrutinize the most frequent non-stop words in the captions to assess content richness. The top 50 most frequent words are shown in \cref{fig: top_word_freq}. Common but meaningless stop words such as ``the,'' ``is,'' and ``and'' are filtered out. The remaining words are sorted by their frequency in descending order. Words suggesting inferential reasoning like ``likely'' and ``possibly'' are prevalent, reflecting the LLM's inference from the OSM tags and its knowledge. For example, in \cref{fig:dataset_inspection}(a), the OSM tags only indicate the presence of a solar power plant, but the LLM extends the semantics by adding ``likely featuring massive gound-mounted photovoltaic panels'' to the caption, connecting shallow tag information (solar power plant) to deeper meanings (photovoltaic panels). Additionally, words like ``edge'', ``straight'' and ``meters'' appear frequently, demonstrating the richness of shape and geography information in the dataset due to the interpretation of attributes in the captioning tasks. For instance, the caption for \cref{fig:dataset_inspection}(d) precisely describes the river's location (from the left-bottom edge to the right-top edge), shape (twist) and length (456 meters). Another noticeable high-frequency word is ``\replaced{county}{conuty},'' which mainly originates from road elements containing a ``tiger:county'' tag. Incorporating this tag into the captions enhances the semantic richness of the image patches by providing additional geographical context, which could be beneficial for future retrieval tasks.

\begin{figure*}[ht]
     \centering
     \subfloat[\centering]{\includegraphics[width=0.30\textwidth]{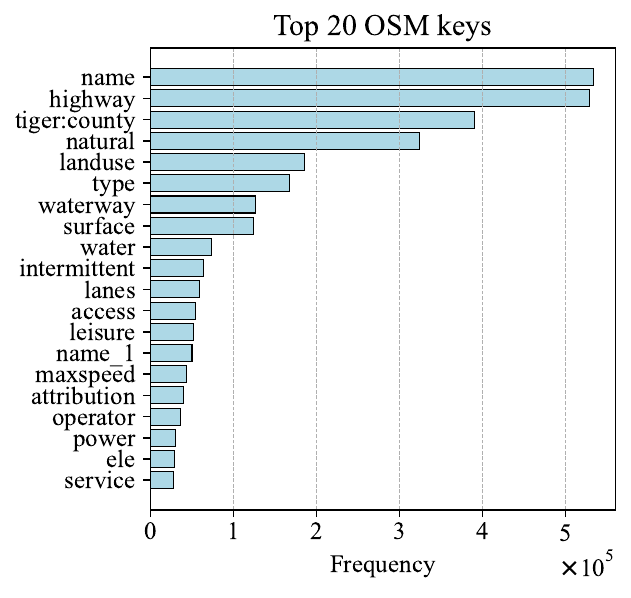}}
     \hfill
     \subfloat[\centering]{\includegraphics[width=0.315\textwidth]{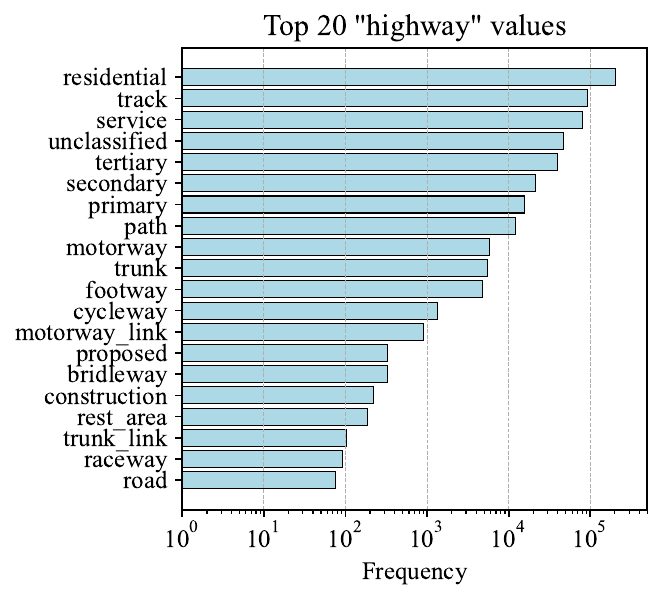}}
     \hfill
     \subfloat[\centering]{\includegraphics[width=0.325\textwidth]{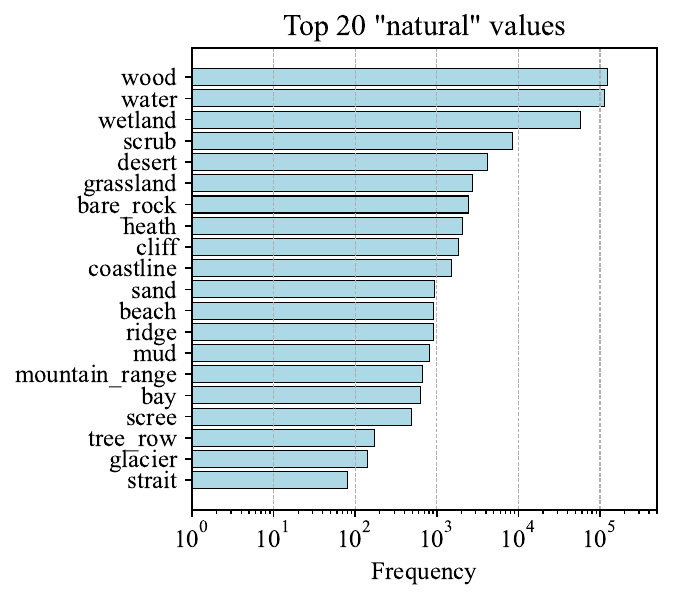}}
     \quad
     \subfloat[\centering]{\includegraphics[width=0.325\textwidth]{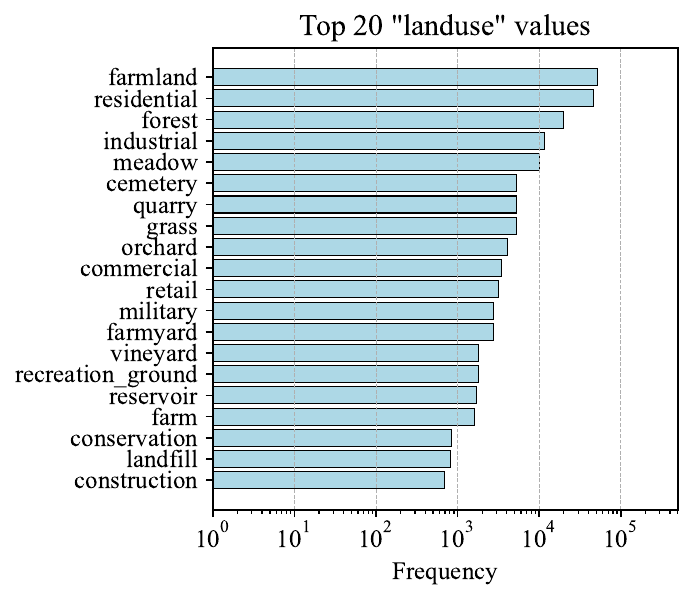}}
     \hfill
     \subfloat[\centering]{\includegraphics[width=0.31\textwidth]{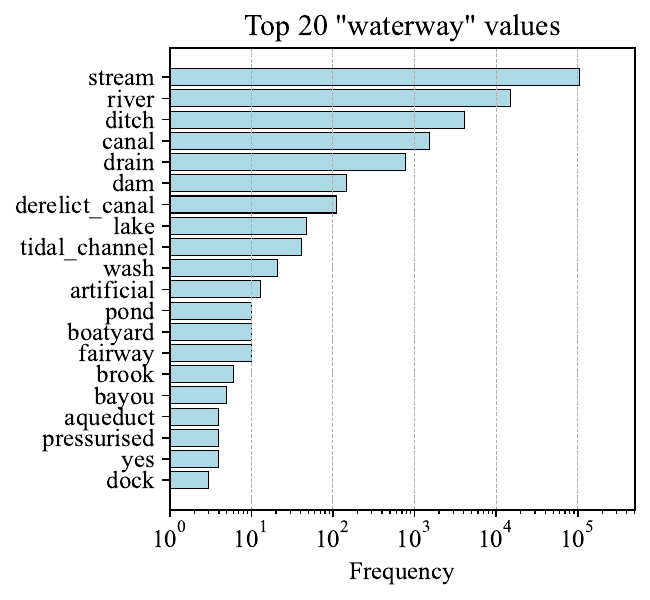}}
     \hfill
     \subfloat[\centering]{\includegraphics[width=0.325\textwidth]{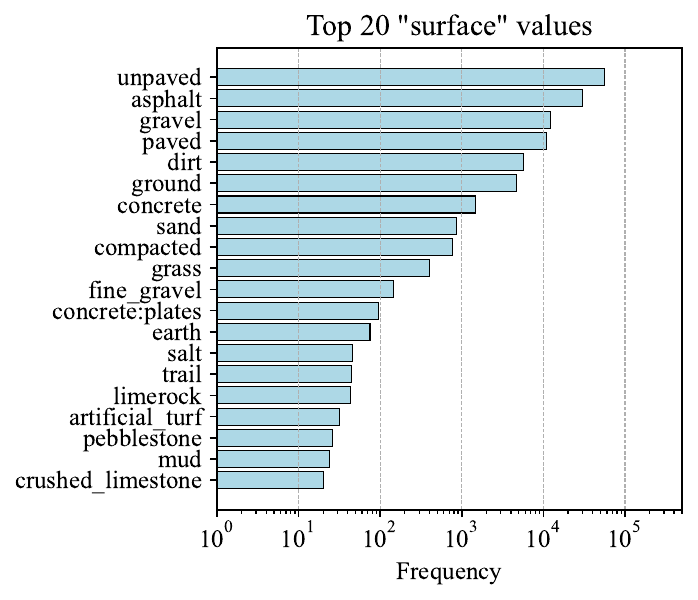}}
     \caption{Topic distribution within the captions of the RSTeller dataset. (a) Top 20 most common OSM keys used in captions. (b-f) Top 20 most frequent values for selected OSM keys used in captioning.}
     \label{fig: caption topic analysis}
\end{figure*}

Additionally, we delve into the topical diversity of the captions by investigating the frequency of the top 20 keys and their corresponding values within the OSM tags used for captioning all the image patches. As shown in \cref{fig: caption topic analysis}(a), the top 20 most frequent keys include feature-related keys such as ``highway'', ``natural'' and ``landuse'', while others are more general like ``name'' and ``tiger:county'' which are usually unique identifiers. To further explore the attribute distribution of some features, we survey the top 20 most frequent values from keys highly related to image features. \cref{fig: caption topic analysis}(b-f) display these for ``highway,'' ``natural,'' ``landuse,'' ``waterway,'' and ``surface''. All values exhibit long-tailed distributions, with significant differences in frequency. The richness of the tag variety ensures the diversity in the topics of all the captions.

Finally, we assess the overall semantic richness of the captions by calculating the MTLD score \citep{mccarthy2010mtld}, a widely recognized metric for evaluating linguistic diversity \citep{ding2023enhancing, herbold2023large, bain2021frozen}, and compare the results with previous works. The MTLD score is computed by treating all captions in the dataset as a single document. To minimize bias introduced by multiple captions for the same image, the captions are concatenated in a random order before calculating the MTLD score for the resulting string. The MTLD score is defined as the mean length of sequential word segments in the text that maintain a minimum threshold Type-Token Ratio (TTR), where TTR represents the ratio of unique words (types) to the total number of words (tokens) in the text, denoted as $\text{TTR}_{\text{thr}}$. This process is mathematically expressed in \cref{eq:mtld}.
\begin{equation}
     \text{MTLD} = \frac{2 \times \text{Total Words}}{\text{Factor}_{\text{fwd}} + \text{Factor}_{\text{bwd}}},
     \label{eq:mtld}
\end{equation}
where $\text{Total Words}$ denotes the total token count of the concatenated text, and $\text{Factor}$ represents the number of segments where each segment maintains a TTR no lower than the predefined threshold, computed both forward ($\text{Factor}_{\text{fwd}}$) and backward ($\text{Factor}_{\text{bwd}}$).

Segmentation begins with the first word, and the cumulative TTR is computed as new words are added. The process halts when the TTR drops below the threshold, forming a segment. The process then restarts from the next word, iterating until the entire text is processed. If a segment is incomplete at the end of the text, a partial factor is computed as $\frac{1 - \text{TTR}}{1 - \text{TTR}_{\text{thr}}}$. For this evaluation, a default TTR threshold of 0.72 is used. The MTLD score quantifies lexical diversity by assessing how consistently unique words are introduced throughout the dataset. Higher MTLD scores indicate greater lexical diversity, reflecting richer vocabulary usage and less repetitive patterns across captions. The results are illustrated in \cref{fig:dataset_comparison}.

As shown in \cref{fig:dataset_comparison}, RSTeller achieves an MTLD score exceeding 100, indicating the highest semantic richness among the datasets. In comparison, earlier datasets such as Sydney-Captions, RSICD, and NWPU-Captions exhibit lower MTLD scores and much smaller dataset scales, underscoring their limitations in both semantic richness and scale. More recent datasets, such as RS5M and SkyScript, achieve moderate improvements in scale but fail to match the semantic diversity and vocabulary richness of RSTeller. Notably, SkyScript has the lowest MTLD score due to its rigid heuristic approach for generating captions. All captions in SkyScript follow a repetitive pattern, such as ``a satellite image of XX of XX," which significantly limits the linguistic diversity and is a major factor contributing to its low score.

In conclusion, the captions in the RSTeller dataset are not only rich and diverse, capturing a wide range of remote sensing features, but also substantial in scale. While these captions include ambiguities and inferred content that may introduce noise, the richness, diversity, and scale of the data are critical for training advanced VLMs in the remote sensing domain. This is particularly true during the pre-training stage, where large-scale data is essential for acquiring foundational vision-language knowledge, and minor levels of noise can be tolerated. However, careful management of these attributes is necessary to minimize potential biases and ensure the reliability and effectiveness of the trained models.

\section{Experiments}
\label{sec:experiments}



To validate the proposed automated dataset generation workflow and assess the effectiveness of the RSTeller dataset, we performed a comprehensive series of experiments. These experiments involved the continual pre-training of the CLIP model with the RSTeller dataset, followed by evaluations of its performance on zero-shot image classification and retrieval tasks. The purpose of these experiments was to rigorously test several hypotheses regarding the utility and impact of the dataset on the training of robust VLMs tailored for remote sensing applications.

The experimental series is structured as follows:

\begin{enumerate}

     \item \textbf{Dataset Effectiveness Evaluation:} We begin by evaluating the effectiveness of the RSTeller dataset through continual pre-training with various CLIP model checkpoints. This step assesses whether incremental learning with our dataset can enhance the model's performance on standard benchmarks. Additionally, this experiment explores the impact of different pre-trained checkpoints on the model's performance.

     \item \textbf{Ablation Study on Training Data:} To assess the contribution of each component in the training data used during the continual pre-training process, we conducted an ablation study, systematically removing one component of the training data at a time. We examined the impact of excluding common knowledge, removing named entities, omitting LLM interpretation, and excluding the caption revision task. This analysis helps identify the significance of each element in enhancing model performance.

     \item \textbf{Data Scalability and Model Performance Analysis:} The final experiment explores the relationship between the performance of VLMs in remote sensing and the scale of domain-specific data, specifically examining the influence of the number of RS-related image-text pairs. This experiment aims to understand how data volume affects model accuracy and generalization capabilities in open-set conditions.
\end{enumerate}

\subsection{Data}

The training data utilized in our experiments consist of three distinct groups: the RSTeller dataset, the SkyScript \citep{wang2024skyscript} dataset, and the LAION-10M dataset, which is a subset of the LAION-400M \citep{schuhmann2021laion}. The detailed information is provided in Table \ref{tab: datasets for exps} and is explained below.

\begin{table}[htbp]
     \centering
     \small
     \caption{Datasets utilized for training}
     \begin{tabular}{ccc}
       \toprule
       Domain & Dataset & Image-Text Pairs \\
       \midrule
       {\multirow{2}*{RS}} & RSTeller & 2,619,852 \\
             & SkyScript & 2,554,817 \\
       \midrule
       Common & LAION-10M & 10,003,420 \\
       \bottomrule
       \end{tabular}%
     \label{tab: datasets for exps}%
\end{table}%

The RSTeller dataset, which we propose, consists of approximately 2.6 million image-text pairs. There are two captions for one image patch in this dataset. To ensure clarity and prevent redundancy during the training of the CLIP model, each image patch appears only once in a training batch. We implement a sampling strategy where image patches for training batches are selected first, followed by a random selection of captions for these images.

The SkyScript dataset, similarly sourced from OSM and GEE, enriches our training with an additional 2.6 million image-text pairs. Each image in this dataset is uniquely associated with one caption. This dataset is integral to augmenting the RS domain knowledge and is merged with the RSTeller dataset to foster a robust study of the relation between model performance and domain data scale in our last experiment.

Lastly, the LAION-10M dataset represents a selective download from the LAION-400M dataset's metadata. Despite some inaccessible URLs, approximately 10 million image-text pairs were compiled. This dataset functions as a supplementary component in our experiments, providing a diverse common knowledge base that aids in preventing catastrophic forgetting and reduces the risk of the model overfitting to specific domain data. This dataset is used across all the experiments as a regulatory fraction of model training.

For evaluative purposes, we employ a series of zero-shot classification and image retrieval tasks. For zero-shot classification, we use eight widely recognized and comparatively extensive benchmark datasets, including AID \citep{xia2017aid}, EuroSAT \citep{helber2019eurosat}, fMoW \citep{Christie_2018_CVPR}, Million-AID \citep{long2021creating}, PatternNet \citep{ZHOU2018197}, NWPU-RESISC45 (RESISC) \citep{cheng2017remote}, RSI-CB256 (RSI-CB) \citep{s20061594}, and ImageNet-1K \citep{imagenet15russakovsky}. The first seven datasets are from the RS domain, while ImageNet-1K pertains to the general domain. The inclusion of ImageNet-1K allows for an evaluation of the model's general-purpose capabilities both before and after the continual pre-training with domain-specific remote sensing data. For image retrieval evaluation, three RS benchmark datasets are used, including UCM Captions \citep{qu2016deep}, RSICD \citep{lu2017exploring}, and RSITMD \citep{9437331}. Details of these benchmark datasets are presented in Tables \ref{tab: datasets for zero-shot classification} and \ref{tab: datasets for zero-shot retrieval}, following the data usage protocols established in the prior study as SkyScript.

\begin{table}[htbp]
     \centering
     \small
     \caption{Benchmark datasets for zero-shot classification evaluation}
       \begin{tabular}{lccc}
       \toprule
       Dataset & Split & Classes & Images \\
       \midrule
       AID   & test  & 30    & 2,000 \\
       EuroSAT & test  & 10    & 2,700 \\
       fMoW  & val   & 62    & 106,081 \\
       Million-AID & train & 51    & 10,000 \\
       PatternNet & train & 38    & 30,400 \\
       RESISC & train & 45    & 31,500 \\
       RSI-CB & train & 35    & 24,747 \\
       ImageNet-1K & val & 1,000 & 50,000 \\
       \bottomrule
       \end{tabular}%
     \label{tab: datasets for zero-shot classification}%
\end{table}%
   
\begin{table}[htbp]
     \centering
     \small
     \caption{Benchmark datasets for zero-shot image retrieval evaluation}
       \begin{tabular}{lccc}
       \toprule
       Dataset & Split & Captions & Images \\
       \midrule
       UCM Captions & test  & 1,050 & 210 \\
       RSICD & test  & 5,465 & 1,093 \\
       RSITMD & test  & 2,260 & 452 \\
       \bottomrule
       \end{tabular}%
     \label{tab: datasets for zero-shot retrieval}%
\end{table}%


\subsection{CLIP Continual Pre-training}

VLMs are typically pre-trained on a large, predefined dataset and have their parameters fixed upon the completion of pre-training. However, given the dynamic nature of the world, the static pre-training dataset may not adequately encompass the knowledge required for specific downstream tasks. To address this gap, continual pre-training is employed to integrate new knowledge into an existing model by incorporating domain-specific data and adhering to the established pre-training paradigm.

Our study adopts continual pre-training to assess the effectiveness of the proposed dataset, evaluate its individual components, and investigate the relationship between the scale of domain-specific data and performance improvements in the CLIP model. Known challenges associated with training exclusively on new domain data include significant performance degradation due to catastrophic forgetting and potential overfitting. To mitigate these challenges, our experiments incorporate data from the original domains, specifically LAION-10M, throughout the continual pre-training process. We consistently sample one-third of the data from LAION-10M for each training batch while the rest are sampled from the RS domain. This technique is applied across all our experiments to ensure balanced training and maintain model stability. Additionally, the effectiveness of this technique is validated the second experiment.


We evaluated two versions of the CLIP model: ViT-B/32 and ViT-L/14. The ViT-B model uses 12 transformer layers and 12 attention heads, while ViT-L has 24 transformer layers and 16 attention heads. The numbers /32 and /14 refer to the patch sizes, $32 \times 32$ and $14 \times 14$ pixels, respectively, processed by the model tokenizer. We used the standard OpenCLIP implementation \citep{ilharco_gabriel_2021_5143773} and initialized the model with four pre-trained checkpoints: the official OpenAI CLIP pre-trained on WIT \citep{radford2021learning}, models trained on LAION-400M \citep{schuhmann2021laion}, DataComp \citep{NEURIPS2023_56332d41}, and LAION-2B \citep{schuhmann2022laionb}. A comparison of the dataset sizes and model checkpoints used is provided in Table \ref{tab: pretrained datasets}. The model checkpoints are represented by the identifier used in the OpenCLIP repository.

\begin{table}[htbp]
     \centering
     \caption{Pre-training datasets and pre-trained checkpoints}
     \resizebox{0.99\columnwidth}{!}{
          \begin{tabular}{lcll}
               \toprule
               Dataset & Size & ViT-B/32 & ViT-L/14 \\
               \midrule
               WIT   & 0.4B  & openai & openai \\
               LAION-400M & 0.4B  & laion400m\_e32 & laion400m\_e32 \\
               DataComp & 1.4B  & datacomp\_xl\_s13b\_b90k & datacomp\_xl\_s13b\_b90k \\
               LAION-2B & 2.3B  & laion2b\_s34b\_b79k & laion2b\_s32b\_b82k \\
               \bottomrule
     \end{tabular}%
     }
     \label{tab: pretrained datasets}
\end{table}%

Experiments were conducted using the OpenCLIP framework, with automatic mixed-precision (AMP) employed to accelerate training and optimize performance. The standard Adam optimizer was used, and both linear warm-up and cosine annealing learning rate schedules were implemented. In the first two experiments, which explore the effectiveness of the RSTeller dataset and perform ablation studies on the components of the training data (\cref{sec: exp dataset effectiveness} and \cref{sec: exp ablation study}), a moderate training configuration was applied to balance the preservation of the pre-trained model feature space with the incorporation of the training data. A learning rate of $1 \times 10^{-8}$ was adopted, and each experiment consisted of 20,000 steps with random data iterations, using a total batch size of 1,024 distributed across 4 NVIDIA A800 GPUs. For the final experiment (\cref{sec: exp domain scaling}), which investigates the scaling of domain data on model performance, a more aggressive training configuration was used, with a learning rate of $1 \times 10^{-5}$ and a total batch size of 8,192 across 4 NVIDIA A800 GPUs, running for 5,000 steps. Gradient checkpointing was employed to minimize memory usage. In the moderate configuration, the model does not fully overfit to the training data, while in the aggressive configuration, the model is forced to fit more closely to the training data.

\subsection{Dataset Effectiveness}
\label{sec: exp dataset effectiveness}
\begin{table*}[htbp]
     \centering
     \caption{Zero-shot top-1 classification accuracy (\%) on benchmark datasets before and after continual pre-training with RSTeller dataset}
     \resizebox{0.85\textwidth}{!}{
     \begin{tabular}{lccccccccc|c}
          \Xhline{1pt}
          Model & Pre-trained Dataset & AID   & EuroSAT & fMoW  & Million-AID & RESISC & PatternNet & RSI-CB & Average & ImageNet-1K \\
          \Xhline{0.5pt}
          \multirow{12}[8]{*}{ViT-B/32} & LAION 2B & 68.10 & 24.40 & 16.00 & 53.75 & 60.46 & 64.36 & 38.11 & 46.45 & 65.84 \\
               & +RSTeller & 61.50 & 29.56 & 14.79 & 50.98 & 55.16 & \underline{65.71} & \underline{42.78} & 45.78 & 65.30 \\
               & $\Delta$ & \textcolor[rgb]{ 1,  0,  0}{\textbf{-6.60}} & \textcolor[rgb]{ 0,  .439,  .753}{\textbf{5.16}} & \textcolor[rgb]{ 1,  0,  0}{-1.21} & \textcolor[rgb]{ 1,  0,  0}{-2.77} & \textcolor[rgb]{ 1,  0,  0}{-5.30} & \textcolor[rgb]{ 0,  .439,  .753}{1.35} & \textcolor[rgb]{ 0,  .439,  .753}{4.67} & \textcolor[rgb]{ 1,  0,  0}{-0.67} & \textcolor[rgb]{ 1,  0,  0}{-0.53} \\
          \cline{2-11}      & LAION 400M & 60.20 & 29.19 & 13.12 & 43.91 & 52.50 & 51.02 & 31.85 & 40.26 & 59.64 \\
               & +RSTeller & 60.50 & 36.96 & 14.92 & 43.84 & 47.31 & 53.63 & 40.43 & 42.51 & 60.08 \\
               & $\Delta$ & \textcolor[rgb]{ 0,  .439,  .753}{0.30} & \textcolor[rgb]{ 0,  .439,  .753}{7.78} & \textcolor[rgb]{ 0,  .439,  .753}{1.80} & \textcolor[rgb]{ 1,  0,  0}{-0.07} & \textcolor[rgb]{ 1,  0,  0}{\textbf{-5.18}} & \textcolor[rgb]{ 0,  .439,  .753}{2.60} & \textcolor[rgb]{ 0,  .439,  .753}{\textbf{8.58}} & \textcolor[rgb]{ 0,  .439,  .753}{2.26} & \textcolor[rgb]{ 0,  .439,  .753}{0.44} \\
          \cline{2-11}      & WIT   & 57.40 & 25.04 & 15.42 & 51.52 & 54.44 & 52.80 & 29.92 & 40.93 & 63.37 \\
               & +RSTeller & 60.60 & 36.41 & 15.22 & \underline{52.06} & 56.17 & 62.11 & 38.21 & 45.82 & 63.59 \\
               & $\Delta$ & \textcolor[rgb]{ 0,  .439,  .753}{3.20} & \textcolor[rgb]{ 0,  .439,  .753}{\textbf{11.37}} & \textcolor[rgb]{ 1,  0,  0}{\textbf{-0.19}} & \textcolor[rgb]{ 0,  .439,  .753}{0.54} & \textcolor[rgb]{ 0,  .439,  .753}{1.72} & \textcolor[rgb]{ 0,  .439,  .753}{9.31} & \textcolor[rgb]{ 0,  .439,  .753}{8.29} & \textcolor[rgb]{ 0,  .439,  .753}{4.89} & \textcolor[rgb]{ 0,  .439,  .753}{0.22} \\
          \cline{2-11}      & DataComp & 66.10 & 24.15 & 15.47 & 49.47 & 62.56 & 60.14 & 35.24 & 44.73 & 68.98 \\
               & +RSTeller & \underline{63.60} & \underline{38.37} & \underline{16.06} & 48.11 & \underline{58.47} & 64.72 & 41.31 & \underline{47.23} & \underline{68.41} \\
               & $\Delta$ & \textcolor[rgb]{ 1,  0,  0}{-2.50} & \textcolor[rgb]{ 0,  .439,  .753}{\textbf{14.22}} & \textcolor[rgb]{ 0,  .439,  .753}{0.59} & \textcolor[rgb]{ 1,  0,  0}{-1.36} & \textcolor[rgb]{ 1,  0,  0}{\textbf{-4.10}} & \textcolor[rgb]{ 0,  .439,  .753}{4.58} & \textcolor[rgb]{ 0,  .439,  .753}{6.07} & \textcolor[rgb]{ 0,  .439,  .753}{2.50} & \textcolor[rgb]{ 1,  0,  0}{-0.57} \\
          \Xhline{0.5pt}
          \multirow{12}[8]{*}{ViT-L/14} & LAION 2B & 67.80 & 38.04 & 24.39 & 63.02 & 70.42 & 72.79 & 42.92 & 54.20 & 75.24 \\
               & +RSTeller & 66.05 & 44.70 & 22.86 & 61.00 & 69.36 & 75.08 & 46.64 & 55.10 & 74.93 \\
               & $\Delta$ & \textcolor[rgb]{ 1,  0,  0}{-1.75} & \textcolor[rgb]{ 0,  .439,  .753}{\textbf{6.67}} & \textcolor[rgb]{ 1,  0,  0}{-1.52} & \textcolor[rgb]{ 1,  0,  0}{\textbf{-2.02}} & \textcolor[rgb]{ 1,  0,  0}{-1.06} & \textcolor[rgb]{ 0,  .439,  .753}{2.29} & \textcolor[rgb]{ 0,  .439,  .753}{3.72} & \textcolor[rgb]{ 0,  .439,  .753}{0.90} & \textcolor[rgb]{ 1,  0,  0}{-0.31} \\
          \cline{2-11}      & LAION 400M & 66.55 & 33.67 & 19.85 & 59.72 & 66.11 & 67.11 & 43.62 & 50.95 & 72.77 \\
               & +RSTeller & 64.95 & 47.70 & 17.29 & 57.82 & 61.89 & 70.11 & 43.99 & 51.96 & 71.97 \\
               & $\Delta$ & \textcolor[rgb]{ 1,  0,  0}{-1.60} & \textcolor[rgb]{ 0,  .439,  .753}{\textbf{14.04}} & \textcolor[rgb]{ 1,  0,  0}{-2.56} & \textcolor[rgb]{ 1,  0,  0}{-1.90} & \textcolor[rgb]{ 1,  0,  0}{\textbf{-4.22}} & \textcolor[rgb]{ 0,  .439,  .753}{3.00} & \textcolor[rgb]{ 0,  .439,  .753}{0.37} & \textcolor[rgb]{ 0,  .439,  .753}{1.02} & \textcolor[rgb]{ 1,  0,  0}{-0.81} \\
          \cline{2-11}      & WIT   & 69.05 & 41.70 & 25.48 & 57.72 & 66.62 & 71.32 & 43.05 & 53.56 & 75.54 \\
               & +RSTeller & \underline{68.95} & \underline{58.78} & 23.31 & 58.51 & 66.61 & 78.31 & \underline{48.96} & 57.63 & 75.69 \\
               & $\Delta$ & \textcolor[rgb]{ 1,  0,  0}{-0.10} & \textcolor[rgb]{ 0,  .439,  .753}{\textbf{17.07}} & \textcolor[rgb]{ 1,  0,  0}{\textbf{-2.17}} & \textcolor[rgb]{ 0,  .439,  .753}{0.79} & \textcolor[rgb]{ 1,  0,  0}{-0.01} & \textcolor[rgb]{ 0,  .439,  .753}{6.99} & \textcolor[rgb]{ 0,  .439,  .753}{5.91} & \textcolor[rgb]{ 0,  .439,  .753}{4.07} & \textcolor[rgb]{ 0,  .439,  .753}{0.16} \\
          \cline{2-11}      & DataComp & 69.65 & 52.56 & 27.57 & 66.89 & 70.98 & 78.22 & 43.59 & 58.49 & 79.19 \\
               & +RSTeller & 66.75 & 56.41 & \underline{24.42} & \underline{63.97} & \underline{69.62} & \underline{78.85} & 45.75 & \underline{57.97} & \underline{79.10} \\
               & $\Delta$ & \textcolor[rgb]{ 1,  0,  0}{-2.90} & \textcolor[rgb]{ 0,  .439,  .753}{\textbf{3.85}} & \textcolor[rgb]{ 1,  0,  0}{\textbf{-3.15}} & \textcolor[rgb]{ 1,  0,  0}{-2.92} & \textcolor[rgb]{ 1,  0,  0}{-1.36} & \textcolor[rgb]{ 0,  .439,  .753}{0.63} & \textcolor[rgb]{ 0,  .439,  .753}{2.17} & \textcolor[rgb]{ 1,  0,  0}{-0.53} & \textcolor[rgb]{ 1,  0,  0}{-0.09} \\
          \Xhline{1pt}
     \end{tabular}%
     }
     \label{tab: cls improvement on checkpoints}%
\end{table*}%
To assess the benefits of the RSTeller dataset for pre-trained models, a series of continual training experiments were conducted on various CLIP checkpoints, evaluating their zero-shot capabilities before and after the continual pre-training. Notably, in this experiment, LAION-10M was utilized to prevent catastrophic forgetting and overfitting.

The evaluation results are presented in Tables \ref{tab: cls improvement on checkpoints} and \ref{tab: ret improvement on checkpoints}. These tables report the top-1 accuracy for zero-shot classification tasks, as well as the average recall rates at top-1 (R@1), top-5 (R@5), and top-10 (R@10) for zero-shot text-to-image retrieval tasks. For each pre-trained checkpoint, the performance of the baseline checkpoints on the benchmark datasets, the performance after continual pre-training on the RSTeller dataset, and the corresponding improvements (marked as $\Delta$) are shown. To provide a comprehensive overview of the model's adaptability, the results from seven remote sensing datasets are averaged to reflect overall performance changes in the remote sensing domain, while results on ImageNet-1K are separately reported to assess the model's robustness against catastrophic forgetting and overfitting. The positive improvements are highlighted in blue, while the negative changes are highlighted in red. For each benchmark, the largest and least improvements are highlighted in boldface, and the best results within the continual pre-trained models are underlined.

Table \ref{tab: cls improvement on checkpoints} presents the top-1 zero-shot accuracy on various benchmark datasets before and after continual pre-training on the RSTeller dataset. Except for the ViT-B/32 model pre-trained on LAION-2B and the ViT-L/14 model pre-trained on DataComp, which exhibit a minor decline in performance on the remote sensing domain, all other models show improvements in the average results (marked as $\Delta$ values in blue). This suggests that the RSTeller dataset enhances the performance of the VLMs, making it a valuable resource for pre-training VLMs in the remote sensing domain. Additionally, all models exhibit minimal performance changes when evaluated on the ImageNet-1K dataset, with most experiencing a slight decline. However, some models, such as ViT-L/14 and ViT-B/32 trained on WIT, show performance gains on ImageNet-1K. This improvement may be attributed to the inclusion of common-domain data during the training process, which positively impacts generalization.

Despite the overall improvements in the average zero-shot classification results, the degree of enhancement varies across different benchmarks. EuroSAT consistently shows the most significant improvements for almost all models and pre-trained checkpoints, with a maximum increase of $17.07\%$ for the ViT-L/14 model pre-trained on WIT. In contrast, some benchmark datasets such as AID, fMoW, and RESISC experienced notable decreases. However, these decreases were not consistent across all pre-trained checkpoints. For example, the ViT-B/32 model pre-trained on LAION-2B showed a significant decrease in AID by $6.60\%$, whereas the same model pre-trained on WIT saw an increase of $3.20\%$ on the same benchmark. This variability suggests that the RSTeller dataset may favor certain pre-trained checkpoints and tasks.

Continual pre-training from different checkpoints for the same model also produces varying results. For both ViT-B/32 and ViT-L/14 models, checkpoints from DataComp consistently demonstrated advantages over other pre-trained models, ranking highest with respect to the average accuracy across seven benchmarks. For most individual benchmarks, checkpoints from DataComp also achieved many of the top results (values with underlines). These findings indicate that a good starting checkpoint for continual pre-training can be advantageous, and checkpoints trained on DataComp are strong candidates for continual pre-training tasks in the RS domain.

Experiment results on zero-shot image retrieval show similar outcomes. As demonstrated in Table \ref{tab: ret improvement on checkpoints}, most checkpoints benefit from continual pre-training on the RSTeller dataset across the benchmark datasets. Notably, there is a slight difference between classification and retrieval tasks: while the checkpoint from DataComp outperforms others for the ViT-L/14 model, the checkpoint from LAION-2B achieves the best results for the ViT-B/32 model after continual pre-training. This variation highlights the nuanced differences between classification and retrieval tasks.

\begin{table}[ht]
     \centering
     \caption{Zero-shot text-to-image retrieval recall rates (\%) on benchmark datasets before and after continual pre-training with RSTeller dataset}
     \resizebox{\columnwidth}{!}{
          \begin{tabular}{lccccc}
               \toprule
               Model & Pre-trained Dataset & RSICD & RSITMD & UCM-Captions & Average \\
               \midrule
               \multirow{12}[8]{*}{ViT-B/32} & LAION 2B & 16.73 & 27.50 & 57.21 & 33.81 \\
                    & +RSTeller & \underline{16.90} & \underline{27.04} & 52.52 & \underline{32.15} \\
                    & $\Delta$ & \textcolor[rgb]{ 0,  .439,  .753}{\textbf{0.17}} & \textcolor[rgb]{ 1,  0,  0}{-0.46} & \textcolor[rgb]{ 1,  0,  0}{\textbf{-4.69}} & \textcolor[rgb]{ 1,  0,  0}{-1.66} \\
               \cline{2-6}      & LAION 400M & 14.49 & 22.54 & 54.38 & 30.47 \\
                    & +RSTeller & 14.07 & 22.54 & \underline{53.49} & 30.03 \\
                    & $\Delta$ & \textcolor[rgb]{ 1,  0,  0}{-0.43} & \textcolor[rgb]{ 0,  .439,  .753}{\textbf{0.00}} & \textcolor[rgb]{ 1,  0,  0}{\textbf{-0.88}} & \textcolor[rgb]{ 1,  0,  0}{-0.44} \\
               \cline{2-6}      & WIT   & 12.24 & 22.34 & 42.44 & 25.67 \\
                    & +RSTeller & 15.84 & 26.79 & 50.66 & 31.10 \\
                    & $\Delta$ & \textcolor[rgb]{ 0,  .439,  .753}{\textbf{3.59}} & \textcolor[rgb]{ 0,  .439,  .753}{4.45} & \textcolor[rgb]{ 0,  .439,  .753}{\textbf{8.22}} & \textcolor[rgb]{ 0,  .439,  .753}{5.42} \\
               \cline{2-6}      & DataComp & 15.88 & 25.03 & 52.87 & 31.26 \\
                    & +RSTeller & 14.11 & 24.59 & 51.90 & 30.20 \\
                    & $\Delta$ & \textcolor[rgb]{ 1,  0,  0}{\textbf{-1.77}} & \textcolor[rgb]{ 1,  0,  0}{\textbf{-0.44}} & \textcolor[rgb]{ 1,  0,  0}{-0.97} & \textcolor[rgb]{ 1,  0,  0}{-1.06} \\
               \midrule
               \multirow{12}[8]{*}{ViT-L/14} & LAION 2B & 20.34 & 31.81 & 62.95 & 38.37 \\
                    & +RSTeller & 20.66 & \underline{32.91} & 60.92 & 38.16 \\
                    & $\Delta$ & \textcolor[rgb]{ 0,  .439,  .753}{0.32} & \textcolor[rgb]{ 0,  .439,  .753}{\textbf{1.10}} & \textcolor[rgb]{ 1,  0,  0}{\textbf{-2.03}} & \textcolor[rgb]{ 1,  0,  0}{-0.21} \\
               \cline{2-6}      & LAION 400M & 16.18 & 26.60 & 54.47 & 32.41 \\
                    & +RSTeller & 16.49 & 28.52 & 53.76 & 32.92 \\
                    & $\Delta$ & \textcolor[rgb]{ 0,  .439,  .753}{0.32} & \textcolor[rgb]{ 0,  .439,  .753}{\textbf{1.92}} & \textcolor[rgb]{ 1,  0,  0}{\textbf{-0.71}} & \textcolor[rgb]{ 0,  .439,  .753}{0.51} \\
               \cline{2-6}      & WIT   & 13.81 & 24.08 & 56.76 & 31.55 \\
                    & +RSTeller & 17.42 & 28.99 & 57.82 & 34.75 \\
                    & $\Delta$ & \textcolor[rgb]{ 0,  .439,  .753}{3.61} & \textcolor[rgb]{ 0,  .439,  .753}{\textbf{4.91}} & \textcolor[rgb]{ 0,  .439,  .753}{\textbf{1.06}} & \textcolor[rgb]{ 0,  .439,  .753}{3.19} \\
               \cline{2-6}      & DataComp & 21.04 & 31.01 & 59.59 & 37.21 \\
                    & +RSTeller & \underline{21.27} & 31.32 & \underline{61.98} & \underline{38.19} \\
                    & $\Delta$ & \textcolor[rgb]{ 0,  .439,  .753}{\textbf{0.22}} & \textcolor[rgb]{ 0,  .439,  .753}{0.31} & \textcolor[rgb]{ 0,  .439,  .753}{\textbf{2.39}} & \textcolor[rgb]{ 0,  .439,  .753}{0.97} \\
               \bottomrule
               \end{tabular}%
     }
     \label{tab: ret improvement on checkpoints}%
\end{table}%

\begin{table*}[ht]
     \centering
     \caption{Ablation study on zero-shot image classification and text-to-image retrieval tasks.}
     \resizebox{1\textwidth}{!}{
     \begin{tabular}{lcccccccc|c|cccc}
          \Xhline{1pt}
          \multicolumn{1}{c}{\multirow{2}[1]{*}{Settings}} & \multicolumn{9}{c|}{Zero-shot Image Classification}                   & \multicolumn{4}{c}{Zero-shot text-to-image retrieval} \\
          \cline{2-14}      & AID   & EuroSAT & fMoW  & Million-AID & RESISC & PatternNet & RSI-CB & Average & ImageNet-1K & RSICD & RSITMD & UCM-Captions & Average \\
          \Xhline{0.5pt}
          Baseline & 68.95 & 58.78 & 23.31 & 58.51 & 66.61 & 78.31 & 48.96 & 57.63 & 75.69 & 16.56 & 28.66 & 59.50 & 34.91 \\
          w/o LAION10M & -1.30 & \textcolor[rgb]{ 0,  .439,  .753}{0.89} & -0.44 & -0.24 & -1.89 & -1.31 & -2.38 & -0.95 & -0.90 & -0.05 & -0.36 & -0.00 & -0.14 \\
          w/o named entities & -0.00 & \textcolor[rgb]{ 0,  .439,  .753}{0.04} & -0.03 & -0.05 & -0.03 & -0.08 & -0.08 & -0.03 & -0.00 & -0.07 & \textcolor[rgb]{ 0,  .439,  .753}{0.09} & -0.00 & \textcolor[rgb]{ 0,  .439,  .753}{0.01} \\
          w/o LLM interpretation & -1.85 & -3.37 & \textcolor[rgb]{ 0,  .439,  .753}{1.29} & -0.94 & -0.57 & -1.05 & -2.32 & -1.26 & -0.01 & \textcolor[rgb]{ 0,  .439,  .753}{0.42} & \textcolor[rgb]{ 0,  .439,  .753}{0.05} & -1.95 & -0.49 \\
          w/o Task 3 & -0.15 & -0.37 & -0.20 & -0.57 & -0.73 & -0.20 & -0.48 & -0.39 & -0.05 & -0.64 & -0.28 & \textcolor[rgb]{ 0,  .439,  .753}{0.27} & -0.22 \\
          \Xhline{1pt}
          \end{tabular}%
          }
     \label{tab: ablation study}%
\end{table*}%

In conclusion, the RSTeller dataset offers researchers a robust starting point to expand their datasets, facilitating improved training of their own VLMs. The positive impact observed across multiple benchmarks underscores the dataset's potential to enhance model performance in the remote sensing domain, thus making it a valuable asset for future research and practical applications.

\subsection{Ablation Study on Training Data}
\label{sec: exp ablation study}

To assess the necessity and effectiveness of the individual components in generating the RSTeller dataset, we conducted an ablation study. This study revisits the previous experiment with different settings, where we systematically removed one component at a time. The baseline for this experiment is based on the ViT-L/14 model, pre-trained on WIT, followed by continual pre-training on both RSTeller and LAION10M as discussed in \cref{sec: exp dataset effectiveness}. The results of this ablation study are shown in \cref{tab: ablation study}, with the first row displaying the baseline results and the subsequent rows showing the performance differences for each setting. Positive differences are highlighted in blue.

In the first experiment, we examined the effect of excluding the LAION10M dataset during the continual pre-training process. Removing this common knowledge resulted in a performance drop across most benchmarks, not only on ImageNet-1K but also on several remote sensing tasks. This suggests that certain common knowledge, present in LAION10M, contributes to the model's ability to generalize well to new domains. Interestingly, the EuroSAT classification task showed a slight improvement. Upon reviewing the confusion matrices, we found that the performance gain primarily came from improved classification of the ``permanent crop'' and ``residential'' classes. This improvement suggests that there might be some noise within the LAION10M dataset that could negatively impact the model's performance on specific classes in the EuroSAT benchmark.

The RSTeller dataset includes named entities such as ``name'' or ``tiger:county,'' which may not directly correspond to visual features in remote sensing images and could potentially introduce noise. To assess their impact, we removed named entities using the spaCy library \citep{spacy2020}, and retrained the model on this modified dataset. The results revealed a negligible drop in performance, indicating that these entities are not detrimental to model performance. While some named entities, such as ``RE Kent South'' from \cref{fig:dataset_inspection}(a), may appear irrelevant to visual content, others, like ``Stonycreek River'' from \cref{fig:dataset_inspection}(d), provide descriptive terms (e.g., ``river'') closely tied to visual features, contributing valuable contextual information during pretraining. Furthermore, named entities may enable broader applications, such as retrieval tasks where specific names or geo-elements are queried. Given their negligible negative impact and potential utility, we conclude that retaining named entities in the dataset is beneficial.

A key contribution of our work is leveraging LLMs to interpret OSM data related to remote sensing images and generate corresponding captions. However, LLMs may introduce hallucinations when inferring the meaning of OSM tags. To investigate the importance of LLM interpretation, we conducted the third experiment by excluding LLMs from the caption generation process. In this experiment, captions were generated using a predefined template, namely ``A remote sensing image of [tag1]; [tag2]; ...,'' formatted with the same filtered OSM tags as in the RSTeller dataset. The results revealed a significant drop in overall performance, underscoring the necessity of including LLMs in the caption generation process. However, we observed that certain benchmarks, including fMoW in classification and RSICD in retrieval, showed improvements when LLM interpretation was excluded. These datasets contain finer-grained classes or more specific content, which may have been obscured by the LLM interpretation. This finding suggests that further research is needed to refine the process for generating more precise captions.

Finally, we evaluated the impact of Task 3, which involves using LLMs to augment captions by revising them in different styles. In this experiment, we removed the Task 3 captions from the dataset. The results demonstrated a notable decline in performance, highlighting the value of this component in the caption generation process. While the performance gain from Task 3 was not large, the task is relatively simple and resource-efficient, requiring fewer tokens for inference. Additionally, revising captions into various styles makes the data synthesis process more versatile and accessible to researchers, especially when raw remote sensing images or OSM data are not readily available. Thus, Task 3 shows promise as a useful tool for enhancing existing multimodal datasets in the remote sensing domain.

\subsection{Model Performance wrt. Domain Data Scale}
\label{sec: exp domain scaling}

\begin{figure*}[ht]
     \centering
     \includegraphics[width=1\textwidth]{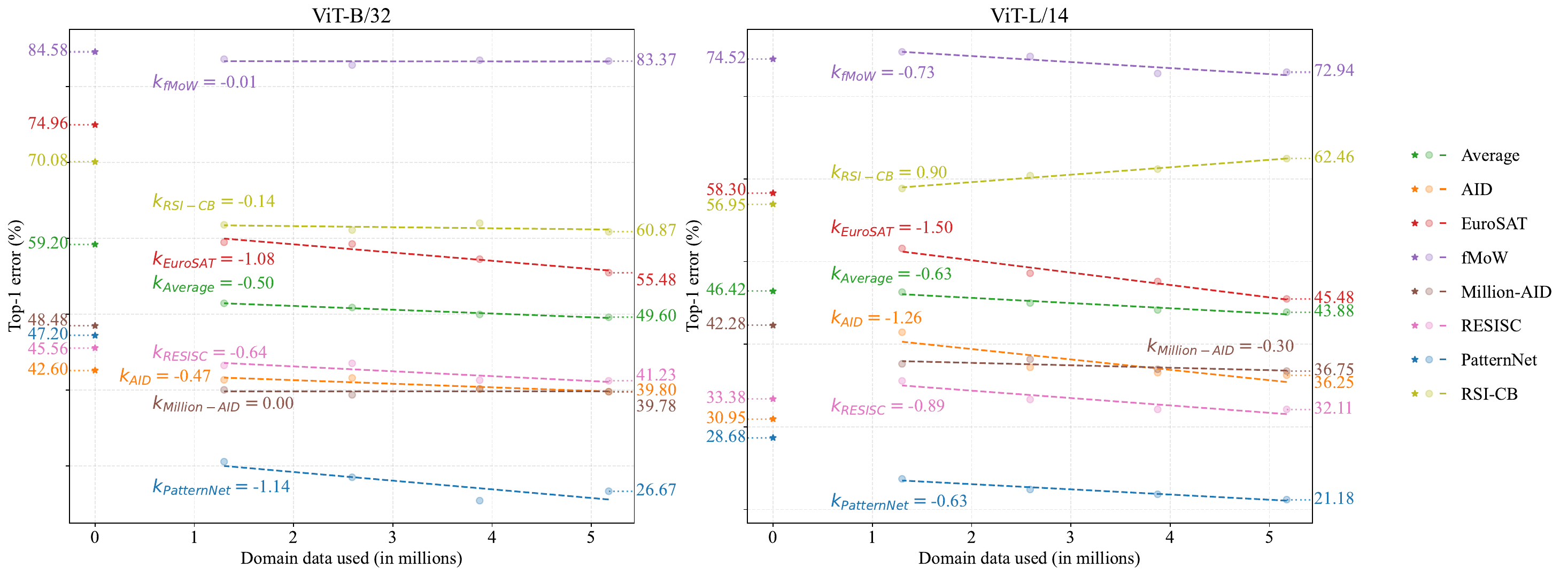}
     \caption{Relationship between domain data scale and zero-shot classification performance. \textbf{Left:} A ViT-B/32 model. \textbf{Right:} A ViT-L/14 CLIP model. Both models are initialized with weight trained on WIT. Scaling domain data consistently improves the performance of the model on the validation dataset.}
     \label{fig: domain scaling classification}
\end{figure*}

\begin{figure*}[ht]
     \centering
     \includegraphics[width=1\textwidth]{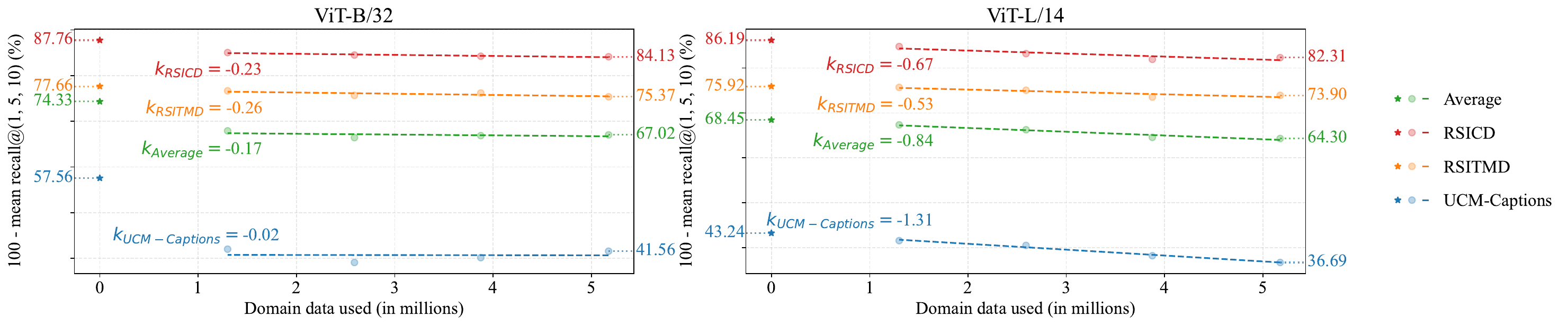}
     \caption{Relationship between domain data scale and zero-shot image retrieval performance. \textbf{Left:} A ViT-B/32 model. \textbf{Right:} A ViT-L/14 CLIP model. Both models are initialized with weight trained on WIT. Scaling domain data consistently improves the performance of the model on the validation dataset.}
     \label{fig: domain scaling retrieval}
\end{figure*}

This experiment was designed to investigate whether scaling the domain-specific data enhances model performance. To this end, we amalgamated our RSTeller dataset with the SkyScript dataset to create a robust RS domain dataset comprising approximately 5.2 million image-text pairs. This extensive dataset allows for a detailed analysis of the model's performance as it relates to variations in domain data volume. We partitioned the data into four equal subsets, each representing about one-fourth of the total data, and ensured no overlap between the subsets. In our experiments, we incrementally incorporated these subsets into the training process, thereby systematically increasing the domain data scale. The ViT-B/32 and ViT-L/14 models pre-trained on WIT were used as the starting point for the continual pre-training. The models were then trained on approximately 25\%, 50\%, 75\%, and 100\% of the domain data, respectively.

The results for zero-shot classification are shown in \cref{fig: domain scaling classification}. The x-axis represents the volume of RS-related domain data used for continual pre-training, while the y-axis indicates the top-1 classification error on the validation dataset. Initially, we evaluated the loaded checkpoints directly on benchmark datasets to establish a baseline for models pre-trained on generic web-based image-text data, represented as stars in the figure. Subsequently, we continually pre-trained the models on RS domain-specific data and evaluated them on the same benchmarks, with the results denoted as dots in the figures. To better visualize performance trends, linear fits were applied to the data points, shown as dashed lines, with the slope values $k$ annotated to reflect the changes in error rates. 

Most benchmarks show improvement when domain data is scaled up, as indicated by the negative slopes of the error rates for both models ($k_{Average}=-0.50$ for ViT-B/32 and $k_{Average}=-0.63$ for ViT-L/14). Datasets such as PatternNet, RESISC, and EuroSAT exhibit substantial improvements, with linear fit slopes of $k_{PatternNet}=-1.14$, $k_{RESISC}=-0.64$ for the ViT-B/32 model, and $k_{EuroSAT}=-1.50$ for the ViT-L/14 model. These results highlight the positive impact of enlarging the domain-specific dataset on the model's classification accuracy. 

However, an unexpected trend is observed with the RSI-CB dataset for the ViT-L/14 model, where top-1 classification error increases as the domain data scale grows ($k_{RSI-CB}=0.90$). Analyzing the model's predictions revealed that certain classes were more frequently misclassified as more RS domain data were incorporated. For instance, ``bare land'' was mistaken for ``desert,'' ``green farmland'' for ``stream,'' and ``mangrove'' or ``sparse forest'' for ``forest,'' while ``resident'' was misclassified as ``avenue.'' These classes often share similar visual features or appear in close proximity within a single image, suggesting potential bias due to granularity and the choice of OSM elements for captioning.

For some benchmarks, such as AID and RESISC evaluated on ViT-L/14, there is an initial performance degradation when domain-specific data is first introduced. In extreme cases, even with all the domain data used for training, the fine-tuned model performance still does not surpass the baseline (e.g., AID for ViT-L/14). Nevertheless, the continual inclusion of domain data leads to consistent performance improvement. For instance, the RESISC benchmark shows a steady decline in error rates despite the initial degradation, as indicated by the slopes ($k_{RESISC}=-0.89$ for ViT-L/14). This trend suggests that further scaling of domain data could potentially surpass the baseline performance of models pre-trained on generic web-based data.

The varied slopes of the linear fits also highlight the differential impact of domain data across different benchmarks. For example, in the ViT-B/32 experiment, the significant negative slope observed for PatternNet ($k_{PatternNet}=-1.14$) compared to the minimal impact on Million-AID ($k_{Million-AID}=-0.00$) suggests that the effectiveness of domain-specific data augmentation varies across different datasets. This variation indicates inherent differences in the benchmarks themselves and possibly the domain specificity of the data they contain.

The experimental results for zero-shot image retrieval tasks, illustrated in \cref{fig: domain scaling retrieval}, reveal trends consistent with those observed in the classification experiments. The x-axis represents the volume of RS-related domain data used for continual pre-training, while the y-axis indicates the evaluation metric, which is the average recall subtracted from 100.

For both the ViT-B/32 and ViT-L/14 models, there is a general improvement in performance with increasing domain data, as indicated by the negative slopes of the linear fits ($k_{Average}=-0.17$ for ViT-B/32 and $k_{Average}=-0.84$ for ViT-L/14). This trend is evident across multiple benchmarks, highlighting the benefits of enlarging domain-specific datasets for image retrieval tasks.

The performance of the ViT-L/14 model, both on classification and retrieval tasks, whether fine-tuned or not, consistently outperforms the ViT-B/32 model in most benchmarks. This is apparent from the lower error rates and higher recall rates across both the original model checkpoints and the models fine-tuned on all the domain data, indicating the superior capability of a larger model in handling various tasks.

The experimental results confirm that increasing domain-specific data generally enhances model performance in the field of remote sensing. Despite the initial degradation observed in specific benchmarks, continuous improvement with increased data volume underscores the potential for further advancements. These findings validate the significance of this work, highlighting the proposed automated workflow's efficiency in generating larger datasets at a low cost. This capability offers substantial benefits to the research community, particularly for training VLMs related to remote sensing.

\section{Discussions}
\label{sec:discussions}

The above experimental results provide several key insights into the impact of scaling domain-specific data on the performance of VLMs as well as the continual pre-training in the remote sensing domain. The findings offer valuable guidance for future research and practical applications in dataset generation and model training.

The experiments in testing the RSTeller dataset (\cref{sec: exp dataset effectiveness}) revealed that the starting checkpoint plays a critical role in the effectiveness of continual pre-training. Models pre-trained on DataComp outperformed those pre-trained on other datasets in most cases, suggesting that selecting a robust initial checkpoint can lead to better results in continual pre-training.

The results in domain data scaling (\cref{sec: exp domain scaling}) confirm that increasing the volume of domain-specific data generally enhances the model performance across various benchmarks. This is evidenced by the consistent improvement in both zero-shot classification and image retrieval tasks as more domain data are included. However, the degree of improvement varies among different benchmarks, indicating that the effectiveness of domain-specific data augmentation is influenced by the nature of the benchmark dataset. For instance, benchmarks like PatternNet and EuroSAT showed significant performance gains, whereas others like RSI-CB exhibited less pronounced improvements or even anomalies in certain cases.

One notable observation is that the initial performance degradation observed in some benchmarks, such as the AID and RESISC, when domain-specific data are first introduced. Despite this initial setback, continuous scaling of the domain data eventually leads to performance improvements, suggesting that the models require a certain threshold of domain-specific data to effectively learn and generalize. This indicates the importance of not being discouraged by the initial performance drops and continuing to expand the dataset for long-term gains.

The ViT-L/14 model consistently outperformed the ViT-B/32 model across most benchmarks, regardless of whether the models were fine-tuned or not. This underscores the advantage of larger models in handling diverse and complex tasks in the RS domain. However, the fine-tuned ViT-L/14 models failed to surpass the original checkpoint performance in some of the benchmarks, even with the full domain data included. This suggests that the overall distribution of the domain dataset created in this experiment could still be different from the benchmark datasets, causing a substantial domain gap. Increasing the volume of data from more diverse sources may bridge this gap.

Based on these findings, several recommendations can be made for future works when trying to bring the VLMs to the RS domain:

\textit{Diverse Data Sources:} When generating RS-domain datasets, it is essential to include data from a variety of sources. This diversity can help the model learn more robust features and improve generalization across different benchmarks. For example, incorporating images with varying ground sample distances, spectral bands, and geographic locations can enhance the dataset's comprehensiveness and utility.

\textit{Threshold Data Volume:} Researchers should aim to exceed a certain threshold of RS-domain data volume to overcome the initial performance degradation and achieve substantial improvements. This highlights the need for large-scale data collection efforts and the importance of not halting the process prematurely.

\textit{Complex Models:} Utilizing more complex models, such as ViT-L/14, can lead to better performance in handling varied tasks within the RS domain. Future research should explore even larger and more sophisticated models that can leverage the enriched domain-specific data effectively.

\textit{Long-Term Data Scaling:} The continuous improvement observed with increasing data volume suggests that long-term efforts in scaling domain-specific datasets will yield better results. Researchers should consider ongoing data collection and integration as a key strategy for advancing the field.

\textit{Selection of Initial Checkpoints:} If choosing to build a VLM from existing pre-trained models, it is essential to identify and utilize robust initial checkpoints that can maximize the benefits of continual pre-training with the RS domain data.


\section{Limitations of the Proposed Work}
\label{sec:limitations}

Despite its strengths, our approach that scales RS domain multimodal data for training RS VLMs has the following limitations that warrant future investigation. 

First, the captioning tasks within our dataset generation workflow are simplistic, typically focusing on a single element, which might not capture all pertinent details in the images. For instance, \cref{fig:dataset_inspection}(b) is captioned solely by its depiction of farmlands, omitting other significant elements like houses and the road nearby. 

Additionally, while our efforts—such as task simplification, tag filtering, wiki-based tag interpretation, and caption refinement—substantially mitigate hallucinations, some residual errors may persist due to the inherent limitations of LLMs. These limitations can occasionally result in speculative or inaccurate outputs. Future work could explore advanced strategies to further suppress hallucinations, including leveraging vision-language models to rate and filter generated captions or integrating uncertainty quantification techniques to flag potentially unreliable outputs.

Moreover, while the RSTeller dataset is a valuable resource, its scale and diversity are not yet comparable to larger datasets such as LAION-400M. It is sourced from a single platform with a fixed ground sample distance and specific image bands, and its geographic coverage is limited to the United States. Subsequent efforts should aim at improving the dataset's quality and expanding its scale to encompass a more diverse array of geographies, image bands and so on. 

Finally, the VLM analyzed in this study, which is a relatively simple CLIP model, does not incorporate recent advancements that would enable more sophisticated interactions with humans in natural language. Future research should explore more complex models that can better understand and utilize the semantics of captions and interact more naturally with human users.

\section{Conclusion}
\label{sec:conclusions}
In this paper, we have meticulously developed and presented an automated dataset generation workflow that enables researchers to efficiently gather and utilize multimodal data for RS applications on a large scale. This workflow optimizes the use of existing, openly available remote-sensing data and employs an LLM to transform raw data into a high-quality, usable dataset for training VLMs in the remote sensing domain. Central to our workflow is the introduction of RSTeller, a multimodal dataset comprising over 2.6 million high-quality image-text pairs with diverse and informative captions, suitable for a wide range of research tasks. Utilizing this dataset for training RS VLMs not only facilitates access to high-quality data for researchers but also reduces the environmental impact associated with generating and maintaining such a dataset. Moreover, we highlight the potential for significant advancements in the field of RS VLMs through the continual enrichment of RS-related multimodal data. We envision future research expanding these datasets to a larger scale, thus enabling the training of more robust VLMs and enhancing the potential prosperity achievable in this field.

\section*{Declaration of Competing Interest}

The authors declare that they have no known competing financial interests or personal relationships that could have appeared to influence the work reported in this paper.

During the preparation of this work the authors used ChatGPT in order to improve readability and language. After using this tool/service, the authors reviewed and edited the content as needed and take full responsibility for the content of the publication.


\section*{Acknowledgments}
This study was supported by the National Natural Science Foundation of China (Grant Nos. 62476205 and 62301405), and the Fundamental Research Funds for the Central Universities (ZYTS25152).

\bibliographystyle{elsarticle-harv} 
\bibliography{reference}

\appendix

\section{Regular Expression for Tag Filtering}
\label{app:regex}

This section presents the regular expression patterns used for tag filtering in the RSTeller dataset.

\begin{lstlisting}[basicstyle=\ttfamily, breaklines=true]
patterns = [
     "^(tiger:(?!seperated|county).*|massgis:.*|nysgissam.*)",
     "^gnis:(f.*|created|.*id|edited)",
     "^(addr:flats|ref.*|FMMP_.*|created_by|source)",
     ".*(wikidata|wikipedia|address|postcode|housenumber|phone|website).*",
     "^[Nn][Hh][Dd](?![:][Ff][Tt][Yy][Pp][Ee]).*"
]
\end{lstlisting}






\end{document}